\documentclass[10pt,twocolumn,letterpaper]{article}

 \usepackage[pagenumbers]{cvpr} %

\definecolor{cvprblue}{rgb}{0.21,0.49,0.74}
\usepackage[pagebackref,breaklinks,colorlinks,allcolors=cvprblue]{hyperref}
\usepackage{xcolor}
\def\confName{CVPR}
\def\confYear{2027}

\newcommand{\ours}{IsoPartFormer}

\title{Keeping Parts Honest: Isolated Part Graphs for Faithful Vision Transformer Explanations}

\author{Bernard Tomczyk \qquad Gail Collyer-Hoar \qquad Bryan Williams\\
Lancaster University\\
Lancaster, United Kingdom\\
{\tt\small \{b.tomczyk, g.collyer-hoar, b.williams6\}@lancaster.ac.uk}
}

\begin{document}
\maketitle
\begin{abstract}
Part and prototype-based explanations for Vision Transformers (ViTs) present a decision as evidence from individual image regions. However, self-attention mixes information across patches, so part-level features absorb context from the whole image and consequently fail to faithfully reflect the decision. We show that existing interpretable ViT classifiers depend heavily on this leakage: when each explained region is encoded in isolation, their accuracy collapses, and retraining them under isolation recovers only part of the lost accuracy. We propose a classifier that represents each image as a graph of isolated parts. Each part is encoded only from its own patches by a ViT whose last blocks are fine-tuned under isolation, preventing evidence leakage from elsewhere in the image. Parts are discovered without annotation and refined hierarchically through their sub-parts, making them more consistent and stable across images than those of competing methods. Each edge is computed from part pairs, and recovers evidence that attention previously shared across the object without affecting the parts. Matching the image graph to a learned prototype graph per class makes the prediction a direct combination of node and edge similarities, each computed only from the patches it is credited to, so every explanation is faithful by construction. Across 11 fine-grained and general-object benchmarks, our method exceeds the accuracy of existing explainable methods that rely on leakage, while providing explanations that reflect what the model actually uses. Code will be made publicly available.
\end{abstract}
    
\section{Introduction}
\label{sec:intro}
\begin{figure}[t]
  \centering
   \includegraphics[width=0.8\linewidth]{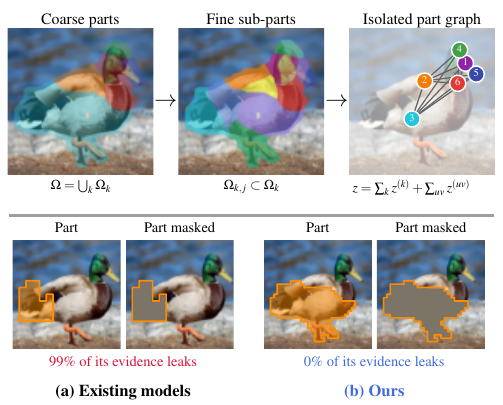}

   \caption{Our approach discovers unannotated coarse and fine sub-parts, encodes each in isolation and connects pairs by edges. The resulting prediction is an exact sum of node $z^{(k)}$ and edge $z^{(uv)}$ terms, each computed only from its credited part or pair (top). When an explained part is replaced by the mean colour, an existing part-based model~\cite{Zhu_2025_CVPR} still credits it 99\% of its original evidence, leaked from elsewhere (a), while ours leaked none (b).}
   \label{fig:teaser}
\end{figure}
Vision Transformers (ViTs)~\cite{dosovitskiy2020image} now underpin much of visual recognition, yet our understanding of how they reach a decision has not kept pace with their accuracy~\cite{suh2025less1percent,KSunnieHelpMe,qiang2025interpretability}. Where decisions must be inspected or contested, as in health~\cite{healthsurvey,health5.0} or autonomous driving~\cite{drivingTrust,DONG2023104358}, an explanation is only useful if it reflects what the model actually used. Post-hoc attributions such as attention rollout and gradient saliency~\cite{abnar2020attentionflow,selvaraju2017gradcam,chefer2021transformerinterpretability} highlight where a model is sensitive, but can be insensitive to the model itself~\cite{adebayo2018sanity}, can be manipulated~\cite{foolingSlack}, and can mislead users into trusting incorrect predictions~\cite{shen2020useful,swamy2025future}. Part- and prototype-based classifiers, instead, build the explanation into the decision itself, classifying an image because a part of it `looks like' a prototypical part~\cite{chen2019protopnet} and reporting the evidence of each part alongside the prediction~\cite{nauta2021prototree,ma2024protovit, ProtoPFormer,PDiscoFormer,Zhu_2025_CVPR}.

Self-attention mixes information across all patches so, after a few blocks, a token located on a part also carries evidence from the rest of the image~\cite{abnar2020attentionflow, aniraj2026metonymyvisionmodelsundermines}, and evidence credited to a part may come from elsewhere, in which case the explanation shows where the model points rather than what it used~\cite{wu2024faithfulness,huang2023evaluation}. We measure for the first time how much existing interpretable ViT classifiers depend on such \textbf{leakage}, by encoding their explained regions in isolation, so that each region sees only its own patches. We find that their accuracy collapses, dropping on CUB-200 \cite{WahCUB_200_2011} by 33 points for PDiscoFormer~\cite{PDiscoFormer} and Zhu~\etal~\cite{Zhu_2025_CVPR}, by 40--48 points for ProtoPNet and ProtoPFormer~\cite{chen2019protopnet,ProtoPFormer}, and by 82 points for ProtoViT~\cite{ma2024protovit}. Retraining them under isolation recovers only part of this and, measured from pixels, 8--52\% of the evidence a typical method assigns to a part changes when the other parts are removed. These classifiers depend on evidence their explanations do not show.

We propose \ours, a classifier that represents each image as a graph of isolated parts (\cref{fig:teaser}) and is built around a single constraint, which we call \emph{region faithfulness}, that every term of the decision is computed only from the pixels to which it is credited. Each part is encoded only from its own patches, with positions expressed relative to the part, by a ViT whose early blocks are frozen and whose last blocks are fine-tuned under isolation. Since an isolated part can no longer borrow context, it must be informative on its own, so we discover parts without annotation by learning coarse object parts jointly with their finer sub-parts, which aligns the sub-parts markedly better with object anatomy. Rather than forbidding interaction between parts, we make it explicit. Two parts can exchange information only within an edge computed from that pair alone, whose term is credited to the pair and leaves the term of each part unchanged. Matching the image graph to a learned prototype graph per class then makes the prediction an exact sum of node and edge terms. Models that are faithful by construction have so far decomposed the decision over pixels~\cite{Boehle2022CVPR}, patches~\cite{HIT}, or feature groups selected per image~\cite{you2025sumofparts}, none of which correspond to the same region across images, whereas our unit is a semantic part shared across images, so that the part-level explanations through which part- and prototype-based models are read become faithful themselves. \textbf{Our key contributions are:}

\begin{enumerate}
    \item We introduce an isolation audit, applicable to any classifier that explains its decision through image regions, and use it to show that state-of-the-art part- and prototype-based ViT classifiers depend heavily on evidence leaked through self-attention, losing most of their accuracy when their explained regions are encoded in isolation.
    \item We propose \ours, a classifier that is region-faithful by construction, in which each part is encoded as a separate token stream with part-relative positions and all interactions between parts are confined to pair edges that are themselves explained terms.
    \item We introduce hierarchical part discovery that learns coarse object parts jointly with nested finer sub-parts without part annotation, yielding more consistent and stable parts across images than competing methods.
    \item We show across 11 benchmarks that this guarantee does not require an accuracy sacrifice, since our model exceeds the mean accuracy of leaky interpretable classifiers and outperforms all of them under the same isolation on 10 of the benchmarks, while its explanations pass the model-randomisation test~\cite{adebayo2018sanity}, show no cross-part conflation, and predict the effect of removing a part far better than competing methods (rank correlation 0.94 vs.\ at most 0.34).
\end{enumerate}

\section{Related work}
\label{sec:related}

\paragraph{Explaining vision transformers.} Post-hoc saliency methods estimate which pixels or regions the output of a trained model is most sensitive to, using gradients, perturbations, class activation maps, or local surrogate models~\cite{selvaraju2017gradcam,sundararajan2017integratedgradients,petsiuk2018rise,ribeiro2016lime,lundberg2017shap}. For Vision Transformers (ViTs)~\cite{dosovitskiy2020image,touvron2021deit}, a further family of methods derives explanations from the attention weights by tracing how information flows between tokens across layers~\cite{abnar2020attentionflow,chefer2021transformerinterpretability}. Such explanations can be insensitive to the model's weights~\cite{adebayo2018sanity}, altered without changing the prediction~\cite{ghorbani2019fragile,foolingSlack}, or unfaithful to the underlying computation~\cite{wu2024faithfulness,swamy2025future}. As such, there is increasing motivation to create models that are interpretable \textit{by design}, in which the explanation is part of the decision process rather than an approximation computed afterwards.

\paragraph{Part- and prototype-based classifiers.} Concept-based methods showed that deep networks can encode human-recognisable concepts~\cite{bau2017networkdissection,kim2018tcav}, with concept bottleneck models making this explicit by routing predictions through a layer of named concepts~\cite{koh2020cbm}, with recent work tying each concept to the attention map of a single object part~\cite{ramachandram2026pfcbm}. Prototype-based classifiers apply the same principle to object parts~\cite{chen2019protopnet}, classifying an image by the similarity of its regions to prototypical parts learned for each class. Subsequent work improved the prototypes themselves, organising them hierarchically~\cite{nauta2021prototree} and making them sparser and more part-like~\cite{nauta2023pipnet}. With the adoption of ViTs, this approach was carried over to transformer backbones, where prototypes have been restricted to local neighbourhoods of tokens~\cite{ProtoPFormer,ma2024protovit}, tied to a fixed set of discovered parts~\cite{PDiscoFormer}, and built non-parametrically on foundation-model features~\cite{Zhu_2025_CVPR}, while other interpretable ViTs explain a decision through the attention of class-specific queries or prompts~\cite{chowdhury2025promptcam}. Evaluations of these models have examined whether prototypes remain consistent across images and stable under perturbation~\cite{huang2023evaluation} and whether users can make effective use of them~\cite{kim2022hive}. They have also shown that the visualised region of a prototype need not reflect what the model uses~\cite{hoffmann2021looks,xudarme2023sanity} and that its activation can change when only pixels outside that region are modified~\cite{sacha2024misalignment}, while synthetic objects with removable parts allow part importances to be checked against ground truth~\cite{hesse2023funnybirds}. Common to all of these models is that the evidence for a part is read from the tokens at the part's location, which implicitly assumes that this evidence originates from the part itself. Since self-attention mixes information across the image, this assumption does not necessarily hold in a ViT, which we test directly in \cref{sec:audit}. \paragraph{Part discovery from foundation-model features.} Self-supervised ViTs provide a strong basis for discovering parts without annotation, since the patch features of DINO and DINOv2~\cite{caron2021dino,oquab2023dinov2} separate object parts and match them across images, a property that has been exploited for training-free co-segmentation and semantic correspondence~\cite{amir2021deep}, unsupervised segmentation~\cite{hamilton2022unsupervised}, object discovery by partitioning a graph of patch tokens~\cite{wang2023tokencut}, and the discovery of concepts that recur across a dataset~\cite{ghorbani2019ace}. Dedicated part-discovery methods even learn a fixed set of parts intended to correspond to the same region in every image of a category~\cite{Hung_2019_CVPR,van_der_Klis_2023_ICCV,PDiscoFormer}. We build on the same features, but a part encoded in isolation has access only to its own pixels, and we therefore learn coarse object parts jointly with the finer sub-parts that compose them.

Relations between parts have also been studied extensively in pattern recognition, from deformable part models~\cite{felzenszwalb2010dpm} and capsule networks~\cite{sabour2017capsules} to relation networks~\cite{santoro2017relationnet}, scene graphs~\cite{johnson2015scenegraphs}, and graph neural networks~\cite{kipf2017gcn,velickovic2018gat}. In fine-grained classification, they have been modelled by relational transformers over discovered parts~\cite{zhao2021partrelational}, graphs of part relationships~\cite{chaudhuri2023TRD}, prototypes of part arrangements~\cite{ayoobi2025protoargnet} and part matching with attention restricted to each part~\cite{black2025partmatching}. Since message passing would reintroduce the mixing that isolation removes, our graph instead confines every interaction between parts to an edge computed from a single pair, so that what two parts provide together is credited to that pair and the term of every individual part is unchanged.

\paragraph{Leakage and faithful-by-construction explanations.} Representations that carry information that their explanations do not reveal have been observed in several settings. Concept bottleneck models have been shown to leak label information through their concept representations~\cite{margeloiu2021concept,mahinpei2021promisespitfallsblackboxconcept,NEURIPS2022_944ecf65}. Attention in transformers distributes information widely enough that a single token can come to reflect much of the input~\cite{abnar2020attentionflow}, and ViT features attributed to one part of an object have been found to absorb information from its other parts, with consequences for attribute prediction~\cite{aniraj2026metonymyvisionmodelsundermines}. Such evidence motivates processing parts in separate streams, but how much the accuracy of existing part- and prototype-based ViT classifiers depends on this leakage has not yet been measured.

A complementary line of work avoids leakage by construction, designing models whose output decomposes exactly into the contributions of their inputs. Such models pair each prediction with explicit concept relevances~\cite{SENN}, classify from local patches with a bounded receptive field~\cite{brendel2019bagnet}, express the logit as an exact sum of per-pixel~\cite{Boehle2022CVPR} or per-patch~\cite{HIT} contributions, or encode learned groups of features separately before summing their scores~\cite{you2025sumofparts}. They share our goal and differ mainly in the unit of explanation, from pixels and patches to feature groups selected per image, none of which refers to the same region across images. Our unit is a semantic part that does, encoded in isolation, with interactions confined to explained pair terms and compared against learned class prototypes. Aniraj~\etal~\cite{aniraj2026metonymyvisionmodelsundermines} also encode parts in separate streams, to remove leakage from attribute predictions, whereas we measure how much existing part-based classifiers depend on leakage and build a classifier whose decision is an exact sum over isolated parts and their pairs.

\section{Method}
\label{sec:method}

\begin{figure*}[t]
  \centering
  \includegraphics[width=0.9\linewidth]{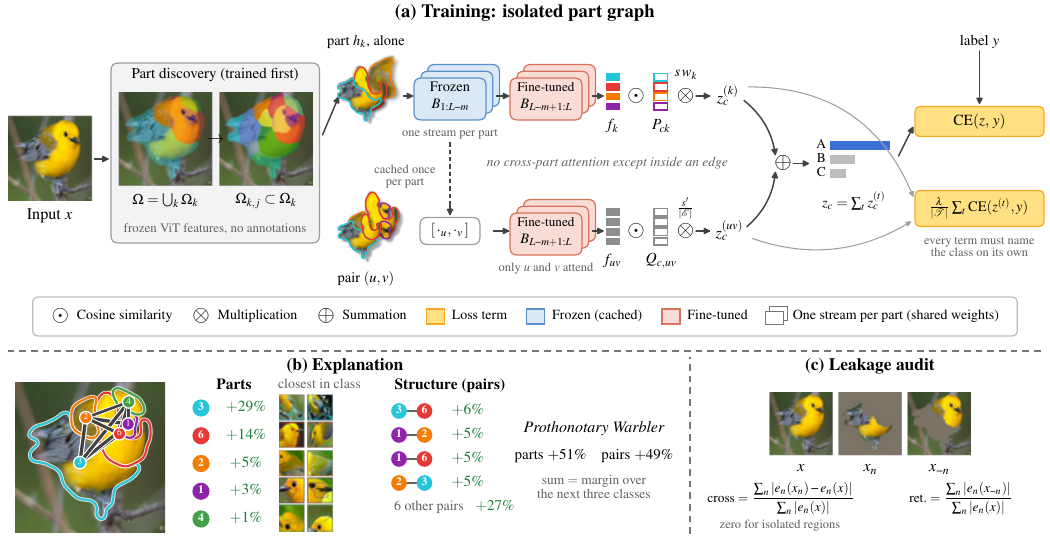}
    \caption{Overview of \ours. (a) Parts discovered without annotations, as coarse parts divided into sub-parts, each pass alone through the frozen ViT blocks (cached once) and the fine-tuned last blocks, giving node descriptors $f_k$, while each edge passes the cached tokens of exactly two parts through the fine-tuned blocks. Matching node and edge descriptors to each class's prototype graph gives terms that sum exactly to the logit $z_c$, and every term is also trained to predict the class on its own. (b) The explanation reads each part's and pair's share of the margin directly from these terms. (c) Our audit compares a region's evidence on $x$, on $x_n$ (all other regions removed) and on $x_{-n}$ (the region itself removed).}

  \label{fig:method}
\end{figure*}

We call an attribution \emph{region-faithful} if the evidence it credits to a region is computed from that region alone. We turn this property into a test for any classifier that explains through regions, apply it to existing classifiers (\cref{sec:audit}), then build a classifier that passes it by construction (\cref{fig:method}) from isolated parts (\cref{sec:discovery,sec:isolation}) whose graph is matched to class prototypes so that the logit decomposes exactly into node and edge terms (\cref{sec:decision}).

\subsection{Isolation and leakage}
\label{sec:audit}

Let a classifier attribute its decision on an image $x$ to regions $R_1,\dots,R_N$, such as the patches of a discovered part or the neighbourhood in which a prototype is matched, and let $e_n(x)$ be the evidence it reports for $R_n$. In a standard ViT every token attends to every patch, so $e_n(x)$ depends on the whole image. We say that $R_n$ is encoded in isolation when its patch tokens, with the class and register tokens, pass through every block without any other patch token present, where other tokens are removed from the sequence rather than masked.

We use isolation to test existing classifiers in two ways, by evaluating a trained classifier with each explained region encoded in isolation, which shows how much of its accuracy depends on leaked information, and by retraining its last blocks and head under isolation, which shows how much of that accuracy its design can recover. To measure leakage directly in pixel space, let $x_n$ be the image with every region other than $R_n$ removed and $x_{-n}$ the image with only $R_n$ removed, where removal replaces pixels by the mean colour, a blurred copy or noise. The cross-part conflation and the retained evidence are
\begin{equation}
  \mathrm{cross} = \frac{\sum_n \lvert e_n(x_n) - e_n(x)\rvert}{\sum_n \lvert e_n(x)\rvert},
  \quad
  \mathrm{ret.} = \frac{\sum_n \lvert e_n(x_{-n})\rvert}{\sum_n \lvert e_n(x)\rvert},
  \label{eq:conflation}
\end{equation}
and cross is zero for any region-faithful classifier.

\subsection{Hierarchical part discovery}
\label{sec:discovery}

An isolated part cannot draw on context to compensate for an inconsistent segmentation, so its evidence is only comparable across images if the part covers the same region of the object in each of them. We discover parts on frozen self-supervised features within a foreground mask $\Omega$ obtained by co-segmentation as in~\cite{amir2021deep}, cropping each image to the bounding box of its mask and using no part or box annotations. Parts form a two-level hierarchy of $K$ coarse parts, each divided into a fixed set of finer sub-parts. Every patch is assigned softly to the coarse parts and a background part by its distance to learned centres in a learned projection of its feature, and the share of each coarse part is divided among its own sub-parts only, so that sub-parts compete with their siblings and lie within their parent (equations in \cref{supp:discovery}).

The assignment is trained with an episodic objective. Each query image is classified by the optimal-transport distance~\cite{pmlr-v97-titouan19a} between its coarse and fine part graphs and class prototypes computed from support images, an objective used only for discovery, together with regularisers for confident and spatially coherent assignments, the presence of every part, decorrelated part descriptors, and a balanced division of every coarse part among its sub-parts. This prevents a single sub-part from absorbing its parent. At test time, each foreground patch is assigned to its most likely coarse part, giving a partition $\Omega = \bigcup_{k=1}^{K}\Omega_k$ shared across images and classes that defines the nodes of the graph.

\subsection{Isolated nodes and edges}
\label{sec:isolation}

Let the ViT have blocks $B_1,\dots,B_L$, of which the last $m$ are fine-tuned and the rest are frozen, and write $B_{a:b}$ for blocks $a$ to $b$ applied in turn. A part $k$ enters the ViT as $h_k$, its own copy of the class and register tokens followed by the patch tokens of $\Omega_k$ only, with position embeddings shifted so that the centroid of the part lies at the centre of the grid. Isolation also removes what attention shared between parts, and rather than forbid such interaction we confine it to edges, each computed from one pair of parts $u,v$ by passing their concatenated frozen intermediate tokens through the fine-tuned blocks. With $\mathrm{avg}$ the mean over patch tokens, the node and edge descriptors are
\begin{equation}
  \begin{aligned}
  f_k &= \mathrm{avg}\, B_{1:L}(h_k), \\
  f_{uv} &= \mathrm{avg}\, B_{L-m+1:L}\bigl([\,B_{1:L-m}(h_u),\, B_{1:L-m}(h_v)\,]\bigr).
  \end{aligned}
  \label{eq:encode}
\end{equation}
A node therefore depends only on the pixels of its part, cross-part attention occurs only in the last $m$ blocks and only between the two parts an edge connects, and the frozen outputs $B_{1:L-m}(h_k)$ are cached once per part, so that isolation holds identically during training and testing, and with $K=6$ parts the graph has at most 15 edges.

\subsection{Prototype graph matching}
\label{sec:decision}

Each class $c$ has a learned prototype $P_{ck}$ for every part and $Q_{c,uv}$ for every pair, and since parts are shared across images, matching requires no search over correspondences. Every node $k$ and present edge $(u,v)\in\mathcal{E}$ forms a term $t\in\mathcal{T}$ whose contribution to the logit of class $c$ is
\begin{equation}
  \begin{aligned}
  z^{(k)}_c &= s\, w_k \bigl(\cos(\bar f_k, P_{ck}) - 1\bigr), \\
  z^{(uv)}_c &= \tfrac{s'}{\lvert\mathcal{E}\rvert} \bigl(\cos(\bar f_{uv}, Q_{c,uv}) - 1\bigr), \quad
  z_c = \textstyle\sum_{t\in\mathcal{T}} z^{(t)}_c ,
  \end{aligned}
  \label{eq:logit}
\end{equation}
where $\bar f$ is a layer-normalised descriptor, $w_k = \lvert\Omega_k\rvert / \lvert\Omega\rvert$ the share of the object covered by part $k$ (zero if absent) and $s, s'$ learned scales. Every term depends only on the patches to which it is credited, so the logit is an exact sum of node and edge contributions and the classifier is region-faithful, with every node credited to its part $\Omega_k$ and every edge to its pair $\Omega_u\cup\Omega_v$. The fine-tuned blocks, prototypes and scales are trained with
\begin{equation}
  \mathcal{L} = \mathrm{CE}(z, y) + \lambda \sum_{\mathcal{S}\in\{\mathcal{V},\mathcal{E}\}} \frac{1}{\lvert\mathcal{S}\rvert} \sum_{t \in \mathcal{S}} \mathrm{CE}\bigl(z^{(t)}, y\bigr),
  \label{eq:loss}
\end{equation}

where $y$ is the label, $\mathcal{V}$ the present parts, and each $z^{(t)}$ omits $w_k$ and $1/\lvert\mathcal{E}\rvert$, so that every term must carry class evidence on its own.

\paragraph{Explanations.} The explanation of a prediction is given by its terms directly. For the predicted class $\hat c$ and the set $\mathcal{N}$ of the next three classes, we report for every term its contrastive contribution $r_t = z^{(t)}_{\hat c} - \frac{1}{3}\sum_{c\in\mathcal{N}} z^{(t)}_c$, so that the contributions sum exactly to the margin of the prediction over those classes, and we show each part next to the same part in training images of the predicted class.

\section{Experiments}
\label{sec:exp}

\subsection{Experimental setup}
\label{sec:exp_setup}

\paragraph{Datasets.} We evaluate on 7 fine-grained benchmarks: CUB-200-2011~\cite{WahCUB_200_2011}, FGVC-Aircraft~\cite{maji13fine-grained}, Stanford Cars~\cite{cars}, Oxford Flowers-102~\cite{Nilsback08}, Oxford-IIIT Pets~\cite{parkhi12a}, NABirds~\cite{NABirds}, Stanford Dogs~\cite{KhoslaYaoJayadevaprakashFeiFei_FGVC2011}, and 4 general-object benchmarks: STL-10~\cite{coates2011stl10}, PASCAL VOC~\cite{Everingham10} object crops, PartImageNet~\cite{he2021partimagenet}, and ImageNet-S50~\cite{gao2022imagenets}. Neither part nor box annotations are used for training or testing, and CUB keypoint annotations are used only to evaluate part quality.

\paragraph{Competitors.} We compare with five part- and prototype-based ViT classifiers~\cite{chen2019protopnet,ProtoPFormer,ma2024protovit,PDiscoFormer,Zhu_2025_CVPR}, with a classifier that encodes groups of features separately by design~\cite{you2025sumofparts} and, for accuracy only, with a classifier whose logit decomposes exactly over pixels~\cite{BCosViT} since its explanations are not organised into parts. All competitors are trained with their official code and backbones, and we refer to~\cite{Zhu_2025_CVPR} as Zhu~\etal. Each is also evaluated with its explained regions encoded in isolation as described in \cref{sec:audit}, first with its trained weights unchanged and then, after retraining, its last blocks and head under isolation, keeping the better of its own training schema and ours on held-out data.

\paragraph{Implementation details.} Our main model uses a DINOv2 ViT-g/14 with registers~\cite{oquab2023dinov2} whose last $m=4$ of its 40 blocks are fine-tuned, and we also report a ViT-B/14 with its last 6 blocks fine-tuned, so that it can be compared with competitors of the same size. Images are cropped to the foreground mask and resized to $448\times448$, and the effect of this crop on our model and on the competitors is reported in \cref{sec:supp_crop}. Part discovery uses $K=6$ coarse and $K_f=16$ fine parts (other values of $K$ in \cref{sec:supp_num_parts}), and all 15 pairs of coarse parts form the edges of the graph. The classifier is trained for 10 epochs with AdamW, learning rate of $4\times10^{-5}$ for the backbone and $10^{-3}$ for the prototypes, weight decay 0.2, batch size 16, and $\lambda=0.5$. The epoch is selected on a held-out 10\% of the training set, predictions are averaged over a horizontal flip, and accuracies of our model are averaged over up to three seeds. The training settings were chosen on a held-out split (\cref{sec:supp_hparams}).

\subsection{Existing part-based ViTs depend on leakage}
\label{sec:exp_leak}

\begin{table*}[t]
\centering
\small
\setlength{\tabcolsep}{4pt}
\caption{Accuracy (\%) with and without leakage. \emph{Leaky}, the original method (for Ours, the same model with full attention); \emph{Isolated}, the same weights with each explained region encoded only from its own patches; \emph{Retrained}, the last blocks and head retrained under isolation. Ours is isolated by construction. Bold, best per column excluding \emph{Isolated}. Other datasets in \cref{sec:supp_leakage_all}.}
\label{tab:leakage}
\begin{tabular}{llccccccccc}
\toprule
& & \multicolumn{3}{c}{CUB} & \multicolumn{3}{c}{Aircraft} & \multicolumn{3}{c}{Cars} \\ \cmidrule(lr){3-5}\cmidrule(lr){6-8}\cmidrule(lr){9-11}
Method & Backbone & Leaky & Isolated & Retrained & Leaky & Isolated & Retrained & Leaky & Isolated & Retrained \\
\midrule
ProtoPNet~\cite{chen2019protopnet} & CaiT-XXS & 82.9 & 42.6 & 54.3 & 84.2 & 33.8 & 56.2 & 90.8 & 42.8 & 60.0 \\
ProtoPFormer~\cite{ProtoPFormer} & CaiT-XXS & 84.0 & 35.7 & 43.1 & 84.3 & 38.8 & 60.2 & 90.8 & 40.4 & 52.0 \\
ProtoViT~\cite{ma2024protovit} & CaiT-XXS & 82.5 & 0.7 & 15.4 & 86.1 & 1.3 & 33.6 & 90.2 & 0.6 & 21.4 \\
PDiscoFormer~\cite{PDiscoFormer} & ViT-B & 88.8 & 55.9 & 85.1 & 50.6 & 29.5 & 84.7 & 85.6 & 30.8 & 88.1 \\
Zhu~\etal~\cite{Zhu_2025_CVPR} & ViT-B & 89.2 & 56.6 & 80.3 & 91.0 & 30.2 & 81.9 & 94.2 & 1.9 & 21.9 \\
SOP~\cite{you2025sumofparts} & ViT-B & 80.5 & 75.5 & 87.9 & 60.4 & 50.0 & 82.1 & 14.6 & 15.2 & 64.2 \\
\midrule
\ours & ViT-B & 92.2 & -- & 90.7 & 93.5 & -- & 90.3 & 94.7 & -- & 92.8 \\
\ours & ViT-g & \textbf{92.6} & -- & \textbf{91.9} & \textbf{94.6} & -- & \textbf{91.5} & \textbf{95.0} & -- & \textbf{93.3} \\
\bottomrule
\end{tabular}
\end{table*}

\begin{table}[t]
\centering
\small
\setlength{\tabcolsep}{3pt}
\caption{Cross-part conflation and retained evidence (\%, CUB, 200 test images, \cref{eq:conflation}). Removed pixels are replaced by mean colour, blur or noise. \ours\space(leaky) is our model with full attention.}
\label{tab:conflation}
\resizebox{\columnwidth}{!}{%
\begin{tabular}{lcccccc}
\toprule
& \multicolumn{2}{c}{Mean colour} & \multicolumn{2}{c}{Blur} & \multicolumn{2}{c}{Noise} \\ \cmidrule(lr){2-3}\cmidrule(lr){4-5}\cmidrule(lr){6-7}
Method & Cross $\downarrow$ & Ret.\ $\downarrow$ & Cross $\downarrow$ & Ret.\ $\downarrow$ & Cross $\downarrow$ & Ret.\ $\downarrow$ \\
\midrule
ProtoPNet & 49.0 & 83.7 & 52.3 & 86.6 & 46.9 & 75.9 \\
ProtoPFormer & 33.9 & 83.7 & 38.5 & 86.0 & 31.3 & 61.6 \\
ProtoViT & 9.8 & 82.5 & 11.2 & 87.5 & 8.8 & 76.5 \\
PDiscoFormer & 42.8 & 66.0 & 21.0 & 108.4 & 42.6 & 62.5 \\
Zhu~\etal & 20.7 & 85.6 & 8.3 & 96.8 & 18.1 & 80.2 \\
SOP & 1.7 & 11.6 & 1.8 & 20.1 & 2.2 & 11.5 \\
\midrule
\ours\space(leaky) & 20.4 & 86.0 & 11.9 & 94.7 & 21.2 & 81.1 \\
\ours & \textbf{0.0} & \textbf{5.9} & \textbf{0.0} & \textbf{18.8} & \textbf{0.0} & \textbf{6.2} \\
\bottomrule
\end{tabular}}
\end{table}

\begin{figure}[t]
  \centering
   \includegraphics[width=0.65\linewidth]{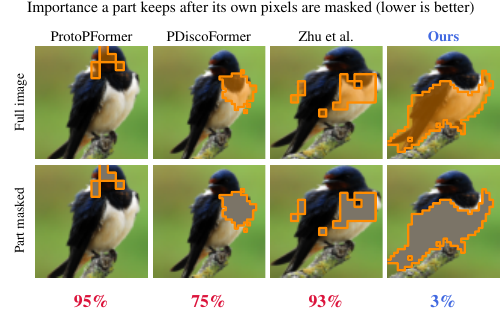}

    \caption{Share of a part's evidence that a model still credits to it after the part's pixels are replaced by the mean colour (top, part; bottom, masked). Evidence that truly comes from the part should vanish when it is masked, so lower is better. Existing models keep 75--95\%, which must come from the rest of the image, whereas ours keeps 3\%, which isolation guarantees comes from the mean-colour region itself rather than from other parts.}

   \label{fig:conflation}
\end{figure}

\Cref{tab:leakage} shows that every part- and prototype-based classifier loses most of its accuracy when its explained regions can no longer exchange information. On CUB, isolation reduces the accuracy of PDiscoFormer and Zhu~\etal by 33 points, that of ProtoPNet and ProtoPFormer by 40 and 48 points, and that of ProtoViT almost to chance, with drops of the same order on Aircraft and Cars. Retraining recovers only part of this, and ProtoViT, ProtoPNet, and ProtoPFormer remain 24 to 69 points below their leaky accuracy. The exception is SOP~\cite{you2025sumofparts}, which already encodes its feature groups separately and even improves when retrained. The collapse of ProtoViT is not an artefact of our protocol, since using our parts as its regions recovers 44.8\% on CUB (\cref{sec:supp_protovit_ourparts}). \ours\space loses 0.7 to 3.2 points relative to the same model with full attention, which is the price of explanations that leak nothing, and remains more accurate than every retrained competitor and than every leaky competitor on CUB and Aircraft.

\Cref{tab:conflation} measures the same dependence in pixel space. For the part- and prototype-based competitors, 8 to 52\% of the evidence credited to a part changes when the other parts are removed, and 62 to 108\% remains after the part's own pixels are removed (\cref{fig:conflation}), so their part evidence is largely carried by the rest of the image. Our model has zero conflation under every form of removal, while its leaky variant is as conflated as the competitors, so the effect is caused by attention rather than by the choice of parts, and it holds for blur and noise as well as a constant colour (\cref{sec:supp_leaky_vs_iso}; examples of every method and form of removal in \cref{sec:supp_vis}).

\subsection{Faithfulness and part quality}
\label{sec:exp_interp}

\begin{table*}[t]
\centering
\small
\setlength{\tabcolsep}{4pt}
\caption{Interpretability on CUB. \emph{Parts}: consistency, stability and distinctiveness of parts following~\cite{Zhu_2025_CVPR,huang2023evaluation} (all test images). \emph{Part-level faithfulness} (200 test images): rank correlation between each part's claimed contribution and the actual effect of removing it, and the gain of deletion and insertion curves ordered by the claimed contributions over a random order (\%). The feature groups of SOP are selected per image and have no identity across images, so part metrics do not apply.}
\label{tab:interp}
\begin{tabular}{lcccccc}
\toprule
& \multicolumn{3}{c}{Parts} & \multicolumn{3}{c}{Part-level faithfulness} \\ \cmidrule(lr){2-4}\cmidrule(lr){5-7}
Method & Consistency $\uparrow$ & Stability $\uparrow$ & Distinctiveness $\uparrow$ & Rank corr.\ $\uparrow$ & Deletion gain $\uparrow$ & Insertion gain $\uparrow$ \\
\midrule
ProtoPNet~\cite{chen2019protopnet} & 2.5 & 54.6 & 64.5 & 0.17 & 10.7 & 9.5 \\
ProtoPFormer~\cite{ProtoPFormer} & 23.4 & 68.0 & 79.2 & 0.14 & 3.2 & 1.4 \\
ProtoViT~\cite{ma2024protovit} & 30.0 & 71.5 & 89.5 & 0.11 & 2.9 & 2.0 \\
PDiscoFormer~\cite{PDiscoFormer} & 49.2 & 70.9 & \textbf{94.8} & 0.34 & 12.0 & 10.4 \\
Zhu~\etal~\cite{Zhu_2025_CVPR} & 63.0 & 79.3 & 93.1 & 0.23 & 9.0 & 9.3 \\
SOP~\cite{you2025sumofparts} & -- & -- & -- & 0.04 & 0.4 & 0.8 \\
\midrule
\ours & \textbf{67.8} & \textbf{93.4} & 92.7 & \textbf{0.94} & \textbf{15.7} & \textbf{16.7} \\
\bottomrule
\end{tabular}
\end{table*}

Our terms sum exactly to the logit, but exactness alone would hold for any linear head, so we test whether a part's claimed contribution predicts the effect of actually removing the part from the image. \Cref{tab:interp} shows that it does, with a rank correlation of 0.94 between claimed and actual effects, against at most 0.34 for the competitors, whose claimed contributions are only weakly related to the effect of removing the corresponding region. The deletion and insertion gains, which reward placing the most influential parts first, are also highest for our model, at 15.7 and 16.7 points against at most 12.0 and 10.4 for the competitors. Pixel-level deletion and insertion metrics for all methods and further details of the randomisation test are given in~\cref{sec:supp_pixel,sec:supp_randomisation}. The explanations also pass the model-randomisation test~\cite{adebayo2018sanity}, as re-initialising the class prototypes, and then also the fine-tuned blocks, reduces the rank correlation between the contrastive contributions of the trained and randomised models to zero on CUB and Aircraft and accuracy to near chance. A swap test confirms isolation directly, since a part replaced by the same part of an image of another class names the original class in at most 0.5\% of cases on the fine-grained datasets and the donor class in 57 to 80\%, while the term of every other part changes by at most 0.044 (\cref{sec:supp_swap}), so an explanation can be edited one part at a time with a predictable effect, which the conflation of leaky classifiers rules out (\cref{tab:conflation}).

Our parts are also the most consistent and the most stable across images among the compared methods, with a consistency of 67.8 and a stability of 93.4 against 63.0 and 79.3 for the strongest competitor, respectively, which indicates that a part of our model refers to the same region of the object across images and under perturbation. The distinctiveness of our parts is close to that of the best method, which learns a dedicated set of discriminative parts~\cite{PDiscoFormer}.

\subsection{Accuracy}
\label{sec:exp_acc}

\begin{table*}[t]
\centering
\small
\setlength{\tabcolsep}{3.5pt}
\caption{Test accuracy (\%) on 11 benchmarks. \textbf{Bold}: best among isolated competitors and Ours; \underline{underlined}: best among leaky competitors and Ours; \textit{italics}: second best among isolated competitors and Ours. NAB: NABirds; VOC: PASCAL VOC object crops; PIN: PartImageNet; IN-S50: ImageNet-S50.}
\label{tab:accuracy}
\begin{tabular}{llcccccccccccc}
\toprule
& & \multicolumn{7}{c}{Fine-grained} & \multicolumn{4}{c}{General objects} \\ \cmidrule(lr){3-9}\cmidrule(lr){10-13}
Method & Backbone & CUB & Aircraft & Cars & Flowers & Pets & NAB & Dogs & STL-10 & VOC & PIN & IN-S50 & Mean\\
\midrule
\multicolumn{13}{l}{\emph{Leaky (original methods)}} \\
ProtoPNet~\cite{chen2019protopnet} & CaiT-XXS & 82.9 & 84.2 & 90.8 & 97.6 & 93.3 & 81.2 & 81.9 & 97.3 & 90.9 & 82.7 & 94.3 & 88.8\\
ProtoPFormer~\cite{ProtoPFormer} & CaiT-XXS & 84.0 & 84.3 & 90.8 & 96.8 & 92.7 & 80.4 & 81.7 & 96.8 & 89.9 & 82.4 & 94.2 & 88.5\\
ProtoViT~\cite{ma2024protovit} & CaiT-XXS & 82.5 & 86.1 & 90.2 & 96.3 & 92.6 & 76.6 & 83.8 & 97.5 & 94.5 & 86.3 & 95.4 & 89.3\\
PDiscoFormer~\cite{PDiscoFormer} & ViT-B & 88.8 & 50.6 & 85.6 & 99.6 & 94.7 & 88.5 & 85.7 & 99.6 & 86.4 & 90.0 & 96.5 & 87.8\\
Zhu~\etal~\cite{Zhu_2025_CVPR} & ViT-B & 89.2 & 91.0 & \underline{94.2} & \underline{99.7} & 95.9 & 88.2 & 88.1 & \underline{99.7} & \underline{96.8} & 88.9 & \underline{98.7} & 93.7\\
SOP~\cite{you2025sumofparts} & ViT-B & 80.5 & 60.4 & 14.6 & 89.4 & 92.8 & 87.6 & 80.4 & 86.2 & 81.4 & 84.9 & 91.5 & 77.2\\
B-cos ViT~\cite{BCosViT} & ViT-B & 75.1 & 59.2 & 72.7 & 90.6 & 90.8 & 67.6 & 89.2 & 97.7 & 88.6 & \underline{90.7} & 93.1 & 83.2\\
\midrule
\multicolumn{13}{l}{\emph{Isolated and retrained}} \\
ProtoPNet & CaiT-XXS & 54.3 & 56.2 & 60.0 & 82.9 & 69.9 & 55.8 & 57.0 & 79.1 & 79.2 & 58.4 & 84.7 & 67.0\\
ProtoPFormer & CaiT-XXS & 43.1 & 60.2 & 52.0 & 73.5 & 53.9 & 68.5 & 49.0 & 72.4 & 75.4 & 69.0 & 67.3 & 62.2\\
ProtoViT & CaiT-XXS & 15.4 & 33.6 & 21.4 & 17.5 & 12.0 & 15.2 & 8.0 & 35.9 & 39.6 & 15.6 & 34.3 & 22.6\\
PDiscoFormer & ViT-B & 85.1 & 84.7 & 88.1 & \textit{99.2} & 91.3 & 83.1 & 75.3 & 96.2 & 86.8 & 74.9 & 93.9 & 87.1\\
Zhu~\etal & ViT-B & 80.3 & 81.9 & 21.9 & 91.1 & 91.5 & 76.7 & 16.2 & \textit{97.9} & \textbf{95.5} & 74.0 & 96.1 & 74.8\\
SOP & ViT-B & 87.9 & 82.1 & 64.2 & 97.2 & \textit{95.2} & 90.5 & 85.3 & 95.3 & 93.4 & 88.1 & 95.0 & 88.6\\
\midrule
\ours & ViT-B & \textit{90.7} & \textit{90.3} & \textit{92.8} & 99.0 & 94.6 & \textit{91.6} & \textit{87.7} & 97.7 & 93.3 & \textit{89.1} & \textit{96.7} & \textit{93.0}\\
\ours & ViT-g & \underline{\textbf{91.9}} & \underline{\textbf{91.5}} & \textbf{93.3} & \textbf{99.6} & \underline{\textbf{96.3}} & \underline{\textbf{93.7}} & \underline{\textbf{90.4}} & \textbf{98.6} & \textit{93.6} & \underline{\textbf{90.7}} & \textbf{97.7} & \underline{\textbf{94.3}}\\
\bottomrule
\end{tabular}%
\end{table*}

\Cref{tab:accuracy} compares accuracy on all eleven benchmarks. Under the same isolation, our model is the most accurate method on every dataset except PASCAL, where the isolated variant of Zhu~\etal remains ahead, and its margins over the best retrained competitor are largest on the most fine-grained benchmarks, at 4.0 points on CUB, 6.8 on Aircraft, and 5.2 on Cars. Against the original competitors, with their leakage, our model is the most accurate on CUB, Aircraft, Pets, NABirds, and Dogs, level with the best on PartImageNet and within 0.1 points of it on Flowers, and behind the best leaky method on Cars, STL-10, PASCAL, and ImageNet-S50. On the general-object benchmarks, part of this gap is explained by our model seeing only the object, whose background a leaky classifier can read, which may matter for classes such as ships and aircraft in STL-10. The ViT-B variant of our model is more accurate than every competitor of the same size on CUB and remains ahead of all isolated competitors on the fine-grained benchmarks except Flowers and Pets, so the advantage of our model does not derive from the larger backbone alone (competitors with a ViT-g backbone are compared in \cref{sec:supp_vitg}).

\subsection{Ablation}
\label{sec:exp_ablation}

\begin{table*}[t]
\centering
\small
\caption{What each component adds (CUB, ViT-g). Each row adds one component to the row above, and bold marks the property it is responsible for. Part quality is NMI / ARI against annotated keypoints at the finest part level, and \emph{Ambiguous} is accuracy on the 25\% of test images on which a separately trained parts-only model is least certain between its two top classes.}
\label{tab:ablation}
\resizebox{\textwidth}{!}{%
\begin{tabular}{lcccccc}
\toprule
& Accuracy $\uparrow$ & Cross-part $\downarrow$ & Faithfulness $\uparrow$ & NMI $\uparrow$ & ARI $\uparrow$ & Ambiguous $\uparrow$ \\
\midrule
Part prototypes (frozen ViT-g, full attention, flat parts) & 89.9 & 8.2 & 0.53 & 60.0 & 31.5 & 68.0 \\
+ isolation & 89.2 & \textbf{0.0 ($-$8.2)} & \textbf{0.85 (+0.32)} & 60.0 & 31.5 & 65.7 \\
+ part refinement (coarse-to-fine) & 89.2 & 0.0 & 0.85 & \textbf{68.6 (+8.6)} & \textbf{51.0 (+19.5)} & 65.4 \\
+ graph (pair edges) & 90.0 & 0.0 & 0.85 & 68.6 & 51.0 & \textbf{68.3 (+2.9)} \\
+ isolated fine-tuning = Ours & \textbf{91.9 (+1.9)} & 0.0 & 0.90 & 68.6 & 51.0 & 74.7 \\
\bottomrule
\end{tabular}}
\end{table*}

\Cref{tab:ablation} builds our model one component at a time, where isolation removes leakage and every later component addresses a cost of isolation. Isolating the parts of a frozen backbone removes all cross-part conflation and raises the faithfulness of the part contributions from 0.53 to 0.85, at a cost of only 0.7 points of accuracy. Refining the parts through their sub-parts leaves accuracy unchanged but brings the sub-parts considerably closer to the annotated anatomy of the birds, raising NMI by 8.6 and ARI by 19.5 points, and the comparison with flat discovery at the same number of parts (60.4 / 35.0) shows that the gain comes from the hierarchy rather than from the larger number of parts (\cref{sec:supp_parts_gt}). The pair edges then add 0.8 points overall and 2.9 points on the quarter of test images in which the appearance of individual parts is most ambiguous. Finally, isolated fine-tuning, which adapts the last blocks to parts and pairs that are each seen alone, adds a further 1.9 points while every term remains computed only from the patches it is credited to. The same fine-tuning applied with full attention would reach 92.6 (\cref{tab:leakage}), but only by teaching the part tokens to draw on the rest of the image, which raises cross-part conflation from 8.2 to 20.4 and lowers faithfulness to 0.31.

\subsection{Qualitative results}
\label{sec:exp_qual}

\begin{figure}[t]
  \centering
  \includegraphics[width=\linewidth]{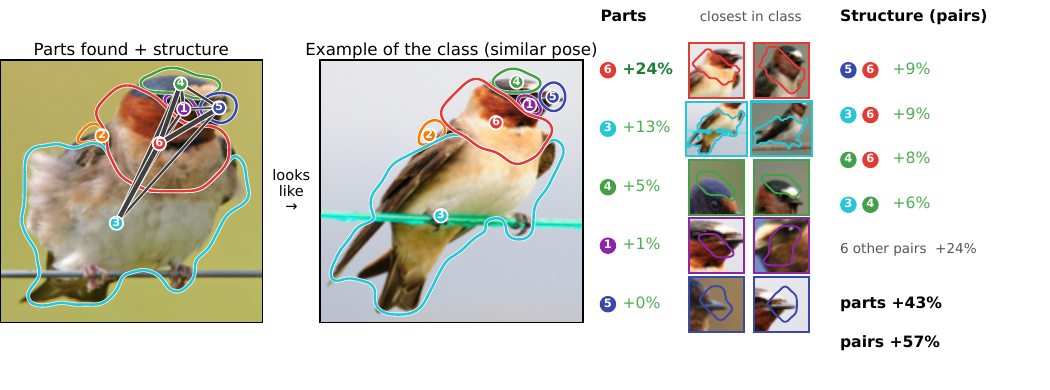}
  \caption{Explanation of our model for a Cliff Swallow (CUB \cite{WahCUB_200_2011}). Left, discovered parts, with pair edges drawn in proportion to their contribution. Middle, training image of the predicted class with a similar part layout. Right, contribution of each part and of the strongest pairs, with each part shown next to the closest same parts in training images of the class.}
  \label{fig:qualitative}
\end{figure}

\Cref{fig:qualitative} shows an explanation of \ours\space for a test image from CUB (further examples in \cref{sec:supp_vis}). Since parts are shared across images, each node can be shown together with the same part in training images of the predicted class, and because the contributions sum to the margin of the prediction, the explanation states precisely how much each part and each pair of parts contributed to choosing one class over its closest alternatives.

\subsection{Preliminary human study}
\label{sec:exp_human}

\begin{table}[t]
\centering
\small
\setlength{\tabcolsep}{3pt}
\caption{Pilot human study on CUB ($N=33$). True class picked among four candidates (all trials / trials the model got right), simulatability, and mean confidence for correct / incorrect answers.}
\label{tab:human}
\resizebox{\columnwidth}{!}{%
\begin{tabular}{lcccc}
\toprule
& \multicolumn{2}{c}{True class picked $\uparrow$} & & \\ \cmidrule(lr){2-3}
Method & All & Model correct & Simulat.\ $\uparrow$ & Confidence \\
\midrule
Grad-CAM~\cite{selvaraju2017gradcam} & 40.4 & 49.0 & 38.4 & 2.70 / 2.34 \\
ProtoViT~\cite{ma2024protovit} & 39.4 & 54.0 & 46.5 & 2.51 / 2.42 \\
PDiscoFormer~\cite{PDiscoFormer} & 25.3 & 30.0 & 33.3 & 2.40 / 2.54 \\
\ours & \textbf{43.4} & \textbf{57.1} & \textbf{51.5} & 2.95 / 2.32 \\
\bottomrule
\end{tabular}}
\end{table}
Fewer than 1\% of explainable AI papers test their explanations with people~\cite{suh2025less1percent}, so we conducted a pilot study with 33 participants following the task-based protocol of~\cite{kim2022hive}, comparing our explanations with those of two part-based classifiers~\cite{ma2024protovit,PDiscoFormer} and a saliency method~\cite{selvaraju2017gradcam} on CUB (Protocol in \cref{sec:supp_human}). \Cref{tab:human} shows that, with our explanations, participants identified the true class most often, predicted the decision of the model most often, and that their confidence separated correct from incorrect answers most clearly, with a significant difference to PDiscoFormer ($p=0.02$). These results suggest that faithful explanations make the decisions of the model easier to anticipate. This research received approval by the Lancaster University FST Ethics Committee.
\section{Conclusion}
\label{sec:conclusion}

We introduced region faithfulness as a testable property of explanations that attribute a decision to image regions, and showed that part- and prototype-based ViTs violate it, losing most of their accuracy once their explained regions are encoded in isolation. \ours\space achieves region faithfulness by construction while achieving the highest mean accuracy among tested interpretable classifiers across 11 benchmarks. Our method's explanations predict the effect of removing a part far better than those of competing methods (rank correlation 0.94 vs.\ at most 0.34), its parts are the most consistent and stable across images, and participants anticipate its decisions most often. The isolation audit needs no part annotations, applies to any classifier that explains its decisions through regions, and exposed leakage in every part- and prototype-based ViT we tested. We recommend it as a standard audit for any such method, and \ours\space shows that it can be passed without sacrificing accuracy.

\section{LLM Usage Statement}
An LLM coding and writing assistant (Claude, Anthropic) was used to help refine text for clarity and length, check citations, debug experiment and write plotting code, and produce figure layouts. All method design, experiments and claims were written and/or verified by the authors, who take full responsibility for the content.
{
    \small
    \bibliographystyle{ieeenat_fullname}
    \bibliography{main}
}

\clearpage
\setcounter{page}{1}
\maketitlesupplementary
\setcounter{section}{6} %
\setcounter{table}{6}   %
\setcounter{figure}{4}  %
\setcounter{equation}{4}%

\section{Part discovery}
\label{sec:supp_discovery}
\subsection{Hierarchical part assignment}
\label{supp:discovery}

Let $q_i$ be a learned linear projection of the frozen key feature of foreground patch $i$, and let $\mu_1,\dots,\mu_K$ be learned coarse-part centres and $\mu_0$ a background centre. The coarse assignment of patch $i$ to part $k$ is
\begin{equation}
  a_{ki} = \frac{\exp\left(-\tau\,\lVert q_i - \mu_k\rVert^2\right)}{\sum_{k'=0}^{K}\exp\left(-\tau\,\lVert q_i - \mu_{k'}\rVert^2\right)},
  \label{eq:supp_coarse}
\end{equation}
with $\tau$ a temperature and $k=0$ the background. Every fine part $j$ has a fixed parent $\pi(j)\in\{1,\dots,K\}$ and a centre $\nu_j$, and receives the share
\begin{equation}
  b_{ji} = a_{\pi(j)i}\;\frac{\exp\left(-\tau\,\lVert q_i - \nu_j\rVert^2\right)}{\sum_{j':\,\pi(j')=\pi(j)}\exp\left(-\tau\,\lVert q_i - \nu_{j'}\rVert^2\right)}
  \label{eq:supp_fine}
\end{equation}
of its parent, a softmax over the children of the same parent only, so that fine parts compete only with their siblings and $\sum_{j:\,\pi(j)=k} b_{ji} = a_{ki}$. At test time patch $i$ is assigned to $\arg\max_{k\ge 1} a_{ki}$ if it lies in the foreground mask, which gives the partition $\Omega=\bigcup_k \Omega_k$ used in the main text. The projection is a linear layer with layer normalisation, each level has its own learned temperature, and patches outside the foreground mask are assigned to the background.

The foreground mask is the co-segmentation of Amir \etal~\cite{amir2021deep}, fitted once on 300 training images with the keys of a frozen DINOv2 ViT-L/14, where a k-means cluster counts as foreground if its [CLS] attention exceeds 0.065 in a large share of these images, with the vote threshold of the original method (\cref{sec:supp_vote_sweep}). The image is cropped to the box of the largest foreground component plus a 5\% margin, and the mask is recomputed in the crop on CUB but carried over from the whole image on the other datasets, where recomputing it keeps too few clusters.

Patches are assigned from the last-block keys of the frozen ViT-g/14, smoothed and reduced to 128 principal components, while the part graphs used during discovery describe each part by its mean output token after a linear discriminant analysis. Each episode of 15 classes with 6 support and 3 query images classifies the queries by the fused Gromov-Wasserstein distance~\cite{pmlr-v97-titouan19a} ($\alpha=0.25$) of their coarse and fine graphs to class prototypes averaged from the support graphs, with the optimal coupling held fixed during back-propagation. Regularisers penalise the entropy and total variation of the assignments and the similarity between part descriptors, and ask every part to appear in the episode, the background to cover the image border and every coarse part to divide its foreground evenly among its children. The network is trained with Adam (learning rate $10^{-3}$, at most 15 epochs of 50 episodes) from k-means centres, with early stopping on 10\% of the training images held out.

\subsection{Foreground mask quality}
\label{sec:supp_mask_quality}
\begin{figure}[]
  \centering
  \includegraphics[width=\linewidth]{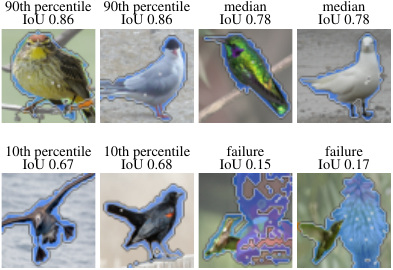}
  \caption{Our foreground mask (white and black outline) against the ground-truth segmentation of CUB (used only for this evaluation), in the cropped view that our model sees, with background inside our mask in blue and bird pixels that our mask misses in orange. Two test images each near the 90th, 50th and 10th percentile of the intersection over union, and two of the worst cases, in which the mask takes in the feeder or the flower that a hummingbird is drinking from.}
  \label{fig:supp_mask_quality}
\end{figure}
Since the parts are restricted to the foreground mask and the image is cropped to it, a poor mask would remove evidence from the bird or let background into the parts. Against the ground-truth segmentations of all 5794 CUB test images, which our pipeline never uses otherwise, the foreground in the crop that the model sees has a mean intersection over union of 0.78 (median 0.79), with a precision of 0.81 and a recall of 0.96, so the mask errs on the side of including a margin around the bird rather than cutting it. The crop keeps 98.4\% of the bird's pixels on average, and only 5.1\% of the images lose more than a tenth of the bird to it. The mask fails, with an intersection over union below 0.5, on 1.1\% of the test images, mostly when another salient object is attached to the bird, as for the hummingbirds at feeders and flowers in \cref{fig:supp_mask_quality}, while two further test images have an empty ground-truth segmentation. The mask uses no information from other test images, since its clusters and their foreground votes are fitted once on 300 training images and every test patch is assigned to its nearest cluster.

\subsection{Datasets}
\label{sec:supp_datasets}
\begin{table}[h]
\centering
\small
\setlength{\tabcolsep}{4pt}
\caption{Datasets. Training and test images as used by every method. Flowers trains on the official training and validation images, PartImageNet tests on its validation and test images together, ImageNet-S50 uses a fixed subset of 400 training images per class and tests on its validation and test images, and PASCAL VOC classifies every non-difficult object of at least 64 pixels on both sides, cropped to its annotated box with a 10\% margin, which defines the task as in prior work. No other box or part annotation is used.}
\label{tab:supp_datasets}
\begin{tabular}{lrrr}
\toprule
Dataset & Classes & Train & Test \\
\midrule
CUB-200-2011 & 200 & 5\,994 & 5\,794 \\
FGVC-Aircraft (variants) & 100 & 6\,667 & 3\,333 \\
Stanford Cars & 196 & 8\,144 & 8\,041 \\
Oxford Flowers-102 & 102 & 2\,040 & 6\,149 \\
Oxford-IIIT Pets & 37 & 3\,680 & 3\,669 \\
NABirds & 555 & 23\,929 & 24\,633 \\
Stanford Dogs & 120 & 12\,000 & 8\,580 \\
\midrule
STL-10 & 10 & 5\,000 & 8\,000 \\
PASCAL VOC 2012 objects & 20 & 9\,398 & 9\,589 \\
PartImageNet & 158 & 20\,466 & 3\,614 \\
ImageNet-S50 & 50 & 20\,000 & 2\,434 \\
\bottomrule
\end{tabular}
\end{table}

\Cref{tab:supp_datasets} lists the eleven datasets with their splits. Every method is trained and tested on the same images, and our pipeline, including the foreground mask and the part discovery, is run with the same settings on every dataset, with no dataset-specific tuning beyond the shared recipe of \cref{sec:supp_hparams}.

\subsection{Vote threshold of the foreground mask}
\label{sec:supp_vote_sweep}
\begin{table}[h]
\centering
\small
\setlength{\tabcolsep}{3.5pt}
\caption{Vote threshold $p$ of the foreground mask: a cluster is foreground if its patches reach the attention threshold in at least $p$\% of the 300 fit images. Agreement of the mask in the model's view with the ground truth (IoU) and share of the object inside the crop (crop recall), on 500 CUB test images (segmentations) and all 752 ImageNet-S50 validation images (segmentations). A dash marks thresholds at which no cluster passes the vote, so that the whole image would be used.}
\label{tab:supp_vote_sweep}
\begin{tabular}{lcccc}
\toprule
& \multicolumn{2}{c}{CUB} & \multicolumn{2}{c}{ImageNet-S50} \\ \cmidrule(lr){2-3}\cmidrule(lr){4-5}
$p$ (\%) & IoU & Crop rec. & IoU & Crop rec. \\
\midrule
5 & 0.15 & 1.00 & 0.31 & 1.00 \\
10 & 0.29 & 1.00 & 0.31 & 1.00 \\
15 & 0.37 & 1.00 & 0.43 & 0.99 \\
20 & 0.53 & 0.99 & 0.59 & 0.95 \\
30 & 0.53 & 0.99 & 0.59 & 0.95 \\
40 & 0.77 & 0.99 & 0.59 & 0.95 \\
50 & 0.77 & 0.99 & 0.59 & 0.95 \\
60 & 0.77 & 0.99 & 0.59 & 0.95 \\
90 & 0.77 & 0.99 & -- & -- \\
95 & 0.38 & 0.99 & -- & -- \\
100 & -- & -- & -- & -- \\
\bottomrule
\end{tabular}
\end{table}

\begin{figure}[]
  \centering
  \includegraphics[width=\linewidth]{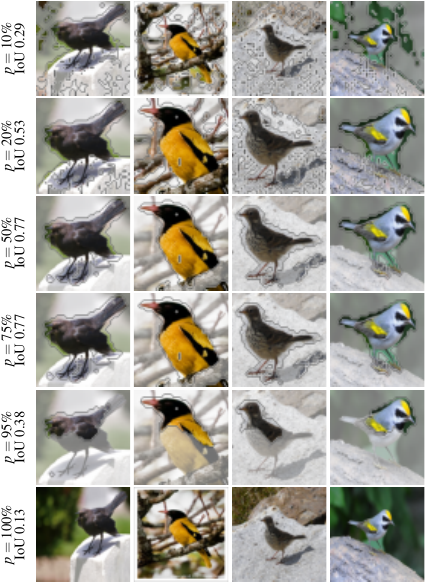}
  \caption{The foreground mask on CUB test images for different vote thresholds $p$, in the cropped view that our model sees, with the mean intersection over union with the ground-truth segmentation over 500 test images. Every threshold from 40 to 90\% selects the same clusters and gives the same mask. Lower thresholds add background clusters, higher ones remove parts of the bird, and at 100\% no cluster passes, so the whole image is used.}
  \label{fig:supp_vote_sweep}
\end{figure}
A cluster of the co-segmentation counts as foreground if its patches are attended by the [CLS] token in at least $p$\% of the fit images. \Cref{tab:supp_vote_sweep,fig:supp_vote_sweep} vary $p$. Since the attention separates the object from the background clearly, the votes of most clusters are either close to all of the fit images or far below half of them, and the mask is the same over a wide range of thresholds, from 40 to 90\% on CUB and from 20 to 60\% on ImageNet-S50. Thresholds below this range add background clusters, and on CUB the agreement with the ground truth falls to 0.53 at 20\% and 0.29 at 10\%, while thresholds above it remove clusters of the object, down to 0.38 at 95\%. The threshold of the original method, which we use, lies inside this range, and the mask is identical at 50\% on CUB, ImageNet-S50, Oxford Pets, Stanford Dogs, NABirds and PartImageNet. On datasets where the object falls into several clusters, such as FGVC-Aircraft, a lower threshold keeps more of the object, and the figures of Aircraft and Cars in \cref{sec:supp_vis} use 50\%.

\subsection{Failure cases of the foreground mask}
\label{sec:supp_mask_failures}
\begin{figure*}[h]
  \centering
  \includegraphics[width=\linewidth]{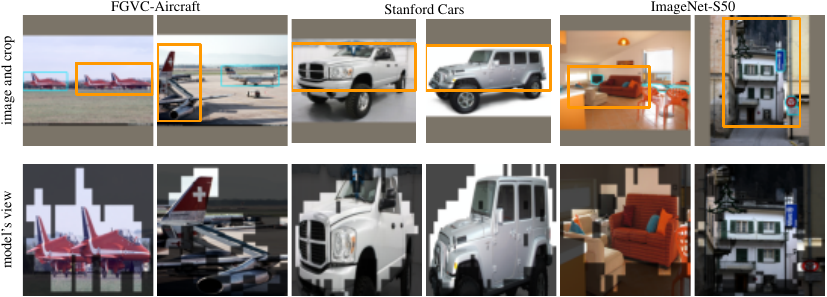}
  \caption{The two worst of 300 random test images for three datasets, with the crop box (orange) and the annotated object (cyan) where an annotation exists, and below the view that our model sees, darkened outside the foreground mask. On Aircraft and ImageNet-S50 the mask selects a different object from the annotated one, here another aircraft, a sofa next to the annotated chair and the house behind the annotated sign. On Stanford Cars, where the wheels and the lower body are dark and close in appearance to the road, the mask can leave them out, so the crop cuts off the tyres.}
  \label{fig:supp_mask_failures}
\end{figure*}
\Cref{fig:supp_mask_failures} shows where the mask fails, with the threshold of 50\% used for the figures of \cref{sec:supp_vis}, chosen as the worst cases among 300 random test images of three datasets, by the share of the annotated aircraft kept in the crop for Aircraft, by the intersection over union with the ground truth for ImageNet-S50, and for Stanford Cars, which has no masks, by how far below the crop the first principal component of the keys still finds the car. The failures are of two kinds. When an image holds several objects of the kind the mask was fitted on, the mask may pick another one than the annotated object, which happens on Aircraft and on ImageNet-S50, whose images often show the annotated object next to larger salient ones (mean intersection over union 0.59 on ImageNet-S50). On Stanford Cars the wheels and the lower body are sometimes not attended by the [CLS] token, so the crop cuts them off. In both cases the model still sees only pixels of the crop and the explanation remains faithful to what it saw, but the evidence of the missing region is lost to it.

\section{Leakage in existing classifiers}
\label{sec:supp_leakage}

\subsection{All datasets}
\label{sec:supp_leakage_all}
\begin{table*}[t]
\centering
\small
\setlength{\tabcolsep}{3.5pt}
\caption{Accuracy (\%) with and without leakage on the remaining eight datasets (as \cref{tab:leakage}). \emph{Isol.}, the trained weights with each explained region encoded in isolation; \emph{Retr.}, retrained under isolation. Ours is isolated by construction; its leaky counterpart (the same model with full attention) was run with ViT-g on all datasets except NABirds and ImageNet-S50.}
\label{tab:supp_leak_other}
\begin{tabular}{lcccccccccccc}
\toprule
& \multicolumn{3}{c}{Flowers} & \multicolumn{3}{c}{Pets} & \multicolumn{3}{c}{NAB} & \multicolumn{3}{c}{Dogs} \\ \cmidrule(lr){2-4}\cmidrule(lr){5-7}\cmidrule(lr){8-10}\cmidrule(lr){11-13}
Method & Leaky & Isol. & Retr. & Leaky & Isol. & Retr. & Leaky & Isol. & Retr. & Leaky & Isol. & Retr. \\
\midrule
ProtoPNet & 97.6 & 65.4 & 82.9 & 93.3 & 60.8 & 69.9 & 81.2 & 43.8 & 55.8 & 81.9 & 42.7 & 57.0 \\
ProtoPFormer & 96.8 & 64.5 & 73.5 & 92.7 & 38.2 & 53.9 & 80.4 & 66.4 & 68.5 & 81.7 & 40.7 & 49.0 \\
ProtoViT & 96.3 & 0.7 & 17.5 & 92.6 & 3.0 & 12.0 & 76.6 & 0.2 & 15.2 & 83.8 & 0.9 & 8.0 \\
PDiscoFormer & 99.6 & 98.0 & 99.2 & 94.7 & 74.7 & 91.3 & 88.5 & 45.6 & 83.1 & 85.7 & 28.6 & 75.3 \\
Zhu et al. & 99.7 & 75.8 & 91.1 & 95.9 & 86.8 & 91.5 & 88.2 & 53.6 & 76.7 & 88.1 & 2.9 & 16.2 \\
SOP & 89.4 & 87.9 & 97.2 & 92.8 & 92.6 & 95.2 & 87.6 & 83.9 & 90.5 & 80.4 & 75.8 & 85.3 \\
Ours (ViT-B) & -- & -- & 99.0 & -- & -- & 94.6 & -- & -- & 91.6 & -- & -- & 87.7 \\
Ours (ViT-g) & 99.7 & -- & 99.6 & 96.0 & -- & 96.3 & -- & -- & 93.7 & 91.9 & -- & 90.4 \\
\midrule
& \multicolumn{3}{c}{STL-10} & \multicolumn{3}{c}{VOC} & \multicolumn{3}{c}{PIN} & \multicolumn{3}{c}{IN-S50} \\ \cmidrule(lr){2-4}\cmidrule(lr){5-7}\cmidrule(lr){8-10}\cmidrule(lr){11-13}
Method & Leaky & Isol. & Retr. & Leaky & Isol. & Retr. & Leaky & Isol. & Retr. & Leaky & Isol. & Retr. \\
\midrule
ProtoPNet & 97.3 & 51.1 & 79.1 & 90.9 & 56.7 & 79.2 & 82.7 & 42.4 & 58.4 & 94.3 & 73.3 & 84.7 \\
ProtoPFormer & 96.8 & 48.0 & 72.4 & 89.9 & 59.6 & 75.4 & 82.4 & 63.3 & 69.0 & 94.2 & 27.2 & 67.3 \\
ProtoViT & 97.5 & 9.6 & 35.9 & 94.5 & 10.6 & 39.6 & 86.3 & 0.7 & 15.6 & 95.4 & 2.5 & 34.3 \\
PDiscoFormer & 99.6 & 86.1 & 96.2 & 86.4 & 67.7 & 86.8 & 90.0 & 55.0 & 74.9 & 96.5 & 85.3 & 93.9 \\
Zhu et al. & 99.7 & 95.0 & 97.9 & 96.8 & 87.4 & 95.5 & 88.9 & 62.2 & 74.0 & 98.7 & 93.5 & 96.1 \\
SOP & 86.2 & 87.2 & 95.3 & 81.4 & 81.2 & 93.4 & 84.9 & 83.6 & 88.1 & 91.5 & 92.6 & 95.0 \\
Ours (ViT-B) & -- & -- & 97.7 & -- & -- & 93.3 & -- & -- & 89.1 & -- & -- & 96.7 \\
Ours (ViT-g) & 99.6 & -- & 98.6 & 95.8 & -- & 93.6 & 91.9 & -- & 90.7 & -- & -- & 97.7 \\
\bottomrule
\end{tabular}
\end{table*}

\Cref{tab:supp_leak_other} repeats Tab.~1 for the eight datasets that do not fit in the main paper, and the pattern is the same on every one of them. Encoding the explained regions in isolation leaves ProtoViT at 0.2 to 10.6\% and costs ProtoPNet and ProtoPFormer between 14 and 67 points, while Zhu \etal\ lose little on STL-10 and ImageNet-S50 but collapse on Stanford Dogs (2.9\%), and PDiscoFormer is barely affected on Flowers but loses 57 points on Dogs. Retraining under isolation recovers part of this loss, and our model is more accurate than every retrained competitor on all eight datasets except PASCAL VOC, where the retrained model of Zhu \etal\ reaches 95.5\% against our 93.6\%. Our own model with full attention, run on six of the eight datasets, is between 0.3 points less and 2.2 points more accurate than our isolated model, with the largest gap on PASCAL VOC, so that isolation costs our model no more here than on the three datasets of Tab.~1 (0.7 to 3.1 points). SOP, which encodes its feature groups separately by design, changes by at most five points under isolation and improves after retraining on every dataset, which shows that our protocol does not penalise a model that does not rely on leaked information.

\subsection{ProtoViT with our parts as its regions}
\label{sec:supp_protovit_ourparts}
\begin{table}[h]
\centering
\small
\setlength{\tabcolsep}{2.5pt}
\caption{ProtoViT on CUB with its own prototype regions or with our parts as the isolated regions (\%). $^\dagger$Released model, trained on ground-truth box crops.}
\label{tab:supp_protovit_ourparts}
\begin{tabular}{lccccc}
\toprule
& & \multicolumn{2}{c}{Own regions} & \multicolumn{2}{c}{Our parts} \\ \cmidrule(lr){3-4}\cmidrule(lr){5-6}
Setting & Leaky & Isol. & Retr. & Isol. & Retr. \\
\midrule
Full image & 82.5 & 0.7 & 15.4 & 8.1 & 44.8 \\
GT box crop$^\dagger$ & 86.6 & 0.6 & 19.9 & 35.0 & 68.6 \\
\bottomrule
\end{tabular}
\end{table}

ProtoViT matches prototypes to small neighbourhoods of a few tokens, so its explained regions cover only 16\% of the image on average (\cref{tab:supp_conflation_ci}), and one may ask whether its collapse under isolation reflects the protocol rather than the model. \Cref{tab:supp_protovit_ourparts} therefore isolates our coarse parts instead of its own regions, so that every isolated stream sees a substantial part of the bird. This raises its retrained accuracy from 15.4 to 44.8\% on full images and from 19.9 to 68.6\% for the released model trained on ground-truth box crops, which confirms that small regions make isolation harder, yet even with our parts it remains 38 and 18 points below its leaky accuracy, so its decision depends on information that its explained regions do not contain.

\subsection{Competitors with a ViT-g backbone}
\label{sec:supp_vitg}
\begin{table*}[t]
\centering
\small
\setlength{\tabcolsep}{3.5pt}
\caption{Competitors with the same DINOv2 ViT-g/14 backbone as our main model (their training settings otherwise unchanged): leaky accuracy and accuracy with the explained regions isolated and the last blocks retrained (\%), for our model the same model with full attention and our isolated model. Bold, best per column. \emph{Black box}, the same backbone fine-tuned as a plain classifier on our crop (the first stage of SOP), with no explanation.}
\label{tab:supp_vitg}
\begin{tabular}{lcccccc}
\toprule
& \multicolumn{2}{c}{CUB} & \multicolumn{2}{c}{Aircraft} & \multicolumn{2}{c}{Cars} \\ \cmidrule(lr){2-3}\cmidrule(lr){4-5}\cmidrule(lr){6-7}
Method & Leaky & Retr. & Leaky & Retr. & Leaky & Retr. \\
\midrule
Black box & 92.1 & -- & 93.3 & -- & 95.1 & -- \\
\midrule
Zhu et al. & 89.5 & 14.1 & 94.1 & 11.8 & \textbf{95.2} & 8.4 \\
PDiscoFormer & 89.2 & 87.7 & 63.4 & 88.1 & 92.8 & 89.0 \\
SOP & 91.3 & \textbf{91.9} & 74.9 & 87.4 & 66.4 & 91.6 \\
\midrule
Ours & \textbf{92.6} & \textbf{91.9}$_{\pm0.3}$ & \textbf{94.6} & \textbf{91.5}$_{\pm0.2}$ & 95.0 & \textbf{93.3}$_{\pm0.2}$ \\
\bottomrule
\end{tabular}
\end{table*}

To separate the effect of isolation from that of the larger backbone, \cref{tab:supp_vitg} trains the competitors with the same DINOv2 ViT-g/14 that our main model uses, with their training settings otherwise unchanged. The larger backbone gives the competitors little with full attention on CUB (Zhu \etal\ 89.5\% against 89.2\% with ViT-B, PDiscoFormer 89.2\% against 88.8\%), and isolation remains far more costly for Zhu \etal\ than for our model, whose isolated accuracy is 0.7 to 3.1 points below its own full-attention counterpart. With a ViT-g backbone, Zhu \etal\ keep their accuracy only when background patches are present, reaching 90.0\% on 300 test images when the whole image is encoded as one isolated region but only 9.3\% when the whole foreground is, against 88.3\% for the same test with ViT-B. The reverse test shows where its evidence lies, since the background encoded alone as one isolated region is classified correctly for 90.0\% of the images with ViT-g but for only 2.7\% with ViT-B, while with full attention and the decision read from the foreground tokens alone both backbones keep about 90\%. The larger backbone therefore places its class evidence in the background tokens, from which the foreground tokens read it through attention, so the explanation credits the bird with evidence that the background supplies, and retraining its last blocks under isolation does not recover from this (14.1, 11.8 and 8.4\%). PDiscoFormer, retrained under isolation in the same way, reaches 87.7, 88.1 and 89.0\%, below our isolated model on all three datasets, and on Aircraft, where its frozen-backbone recipe leaves the leaky model at 63.4\%, retraining its last blocks under isolation raises it to 88.1\%. SOP, which encodes its groups separately by design, does not collapse with ViT-g either and improves when retrained under isolation, as it does with ViT-B, from 91.3 to 91.9\% on CUB, from 74.9 to 87.4\% on Aircraft and from 66.4 to 91.6\% on Cars, so that it matches our isolated model on CUB and remains below it on Aircraft and Cars (91.5 and 93.3\%), while its feature groups overlap almost completely (\cref{sec:supp_conflation}) and therefore do not explain separate regions as our parts do. With full attention, the ViT-g model of Zhu \etal\ is more accurate than our isolated model on Aircraft and Cars, as is our own full-attention counterpart, which is the price of explanations that leak nothing, as discussed in Sec.~4.2. The same backbone fine-tuned as a plain black-box classifier on our crop reaches 92.1, 93.3 and 95.1\% (\cref{tab:supp_vitg}), so our isolated model gives up 0.2 to 1.8 points against a model that explains nothing.

\subsection{Full adaptation under isolation}
\label{sec:supp_full_adaptation}
The retrained competitors of Tab.~1 adapt only their last four blocks and their heads to isolation, so one may ask whether a more thorough adaptation would close the gap. We therefore retrained all twelve blocks of Zhu \etal\ (ViT-B, CUB) under isolation for 30 epochs, with the same learning rates as our model and the epoch selected on held-out training images. It reaches 76.8\%, below adapting only the last four blocks (80.3\%, Tab.~1) and 13.9 points below our isolated ViT-B model (90.7\%), and its held-out accuracy peaks after the first epoch and then declines while its training loss approaches zero, so additional capacity under isolation fits the training images without recovering the evidence that its regions lose.

\subsection{Isolation protocol per method}
\label{sec:supp_iso_protocol}
Every competitor is isolated with its own published regions and its own decision rule, so that only the exchange of information between regions changes. For Zhu \etal, every foreground patch is assigned to the prototype slot of the predicted class that it matches best, the background being the patches where the background prototypes win, and the class score keeps its maximum over all isolated patches of the similarity to each prototype. For PDiscoFormer, the released part maps define the regions, the background is left out, and the mask-weighted part descriptors, their modulation, the class head and the mean over parts are unchanged. ProtoViT, ProtoPFormer and ProtoPNet on a ViT use a class token that attends to every patch, so both channels are isolated, with every region passing through the self-attention blocks alone and receiving its own class token from the class-attention blocks over that region only, while the regions are the patches matched by the prototypes of the predicted class, the global prototypes of ProtoPFormer being matched to the region class tokens. For SOP, every group keeps only its own patch tokens and the prefix tokens in its forward pass, whereas the original method sets the pixels outside the group to zero but still lets the group attend to the resulting tokens, so that the group's position and shape remain visible. In every case each region passes through the model's own backbone with its own patches and the prefix tokens, both the regions and the decision are taken from the model as published, and the regions of the competitors whose regions are defined by the prediction are computed with the unrestricted model, which favours them. Retraining under isolation fine-tunes the last four blocks and every head with flip augmentation, each flip with its own regions, selects the epoch on 10\% of the training images held out, and reads the test set once. Encoding the whole image of Zhu \etal\ as one isolated region reproduces its published accuracy exactly (\cref{sec:supp_collapse}), so the change of attention connectivity costs nothing by itself.

\subsection{Isolation does not cause the collapse}
\label{sec:supp_collapse}
\begin{table}[h]
\centering
\footnotesize
\setlength{\tabcolsep}{2.5pt}
\caption{Controls for the isolation protocol (CUB, all 5794 test images, trained weights unchanged, \%), the values of \cref{fig:supp_collapse}. \emph{Whole image}: every patch encoded as one isolated region, with the model's own decision rule. \emph{Merged}: the explained regions of an image merged by nearest centroids into $k$ isolated regions (only for $k$ below the method's own number of regions). \emph{Own}: every explained region encoded alone, as in Tab.~1 (removed), by removing the other tokens or by a block-diagonal attention mask over one sequence (computed for the methods whose regions have their own prefix tokens). \emph{Regions / image}: mean number of explained regions per image over all test images (on the 200 images of \cref{tab:supp_conflation_ci} PDiscoFormer has 7.9). Published and Own differ from Tab.~1 by at most 0.6 points, since these controls use the best saved checkpoint of ProtoPNet and ProtoPFormer instead of the last epoch and encode every region in single precision through all blocks instead of from the cached half-precision tokens of the first blocks.}
\label{tab:supp_collapse}
\begin{tabular}{lccccc}
\toprule
 & ProtoPNet & ProtoPF. & ProtoViT & PDisco. & Zhu \\
\midrule
Published & 83.5 & 84.3 & 82.5 & 88.9 & 89.2 \\
Whole image & 83.5 & 83.8 & 82.5 & 87.7 & 89.2 \\
\midrule
Merged, $k{=}1$ & 83.5 & 78.3 & 22.3 & 85.5 & 87.8 \\
Merged, $k{=}2$ & 79.4 & 60.9 & 10.4 & 82.4 & 81.4 \\
Merged, $k{=}3$ & 72.9 & 52.4 & 4.5 & 78.0 & 73.9 \\
Merged, $k{=}5$ & 59.1 & 44.3 & 1.4 & 67.4 & -- \\
Merged, $k{=}8$ & -- & 39.1 & 0.6 & -- & -- \\
Own, removed & 42.6 & 35.7 & 0.7 & 56.4 & 56.5 \\
Own, masked & -- & -- & -- & 56.4 & 56.5 \\
\midrule
Regions / image & 6.5 & 9.3 & 9.3 & 7.8 & 5.0 \\
\bottomrule
\end{tabular}
\end{table}

\begin{figure}[h]
  \centering
  \includegraphics[width=\linewidth]{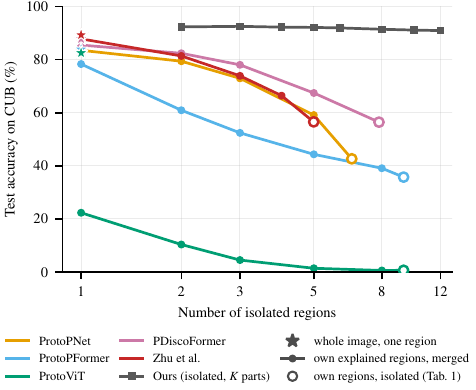}
  \caption{Accuracy of the competitors on CUB as their image is split into more isolated regions, with trained weights unchanged. Lines merge each model's own explained regions step by step and end at its own regions as isolated in Tab.~1 (hollow markers, at their mean number per image), and the stars give the whole image encoded as one isolated region, which reproduces the published accuracy. Our model (grey) is trained in isolation for each number of parts $K$ (\cref{tab:supp_num_parts}), whereas the competitors are evaluated as trained.}
  \label{fig:supp_collapse}
\end{figure}
Three controls show that the collapse of \cref{tab:supp_leak_other} and Tab.~1 comes from the dependence of these classifiers on information exchanged between their regions and not from the isolation protocol (\cref{tab:supp_collapse,fig:supp_collapse}). Encoding every patch of the image as one isolated region, with each model's own decision rule, reproduces its published accuracy (89.2\% for Zhu \etal, 83.5\% for ProtoPNet and 82.5\% for ProtoViT, and within half a point for ProtoPFormer), so taking tokens out of the sequence and passing the rest through the model costs nothing by itself, while PDiscoFormer loses 1.2 points because isolation pools its hard part maps rather than the soft maps of the published model.
Encoding the explained regions in one sequence with a block-diagonal attention mask, instead of removing the other tokens, gives exactly the accuracy of removing them (56.5\% for Zhu \etal\ and 56.4\% for PDiscoFormer), so the collapse does not depend on how isolation is implemented.
Merging the explained regions of every image step by step lowers accuracy steadily with the number of isolated regions, from 87.8 to 56.5\% between one and five regions for Zhu \etal\ and from 83.5\% with one region to 42.6\% with its own regions for ProtoPNet, and splitting the same patches instead into compact chunks by their position, which have nothing to do with the model, falls in the same way (to 64.3\% with five chunks for Zhu \etal\ and to 57.5\% for ProtoPNet, against 59.1\% with five merged regions), so no choice of regions avoids the loss. ProtoViT is the clearest case, since all of its explained regions encoded together as one region reach only 22.3\%, as they cover 16\% of the image while its decision draws on the rest. A model trained on whole images might also lose accuracy on fragments only because it has never seen them, but retraining under isolation recovers only part of the loss (Tab.~1 and \cref{sec:supp_full_adaptation}), SOP, which encodes its groups separately by design, does not collapse, and our model, trained in isolation, stays between 90.9 and 92.5\% for two to twelve parts, so isolated parts can carry the decision when the model is built for them. These controls show that the competitors depend on information exchanged between their regions, while the size of the loss also reflects how far isolated inputs lie from the inputs that a model was trained on, so it should not be read as the share of the decision that leaks.

\section{Faithfulness of our explanations}
\label{sec:supp_faithfulness}
Our guarantee is conditional on the partition shown to the user. Given the parts, every term of the decision is computed only from the pixels of its own part or pair, and all interventions in this section therefore keep the foreground mask and the parts of the original image fixed and recompute only the terms. The contribution we report for a part, its node term plus half of each pair term that contains it, is an additive allocation of the logit that sums exactly to the decision, and \cref{sec:supp_deletion_identity} shows how the exact effect of removing a part follows from the same displayed terms.

\subsection{Cross-part conflation}
\label{sec:supp_conflation}
\begin{table*}[t]
\centering
\small
\setlength{\tabcolsep}{3.5pt}
\caption{Conflation on CUB (mean colour, 200 test images) with 95\% bootstrap intervals, and the explained regions: number per image, mean pairwise IoU between them, and the share of the image they cover together (\%). SOP's feature groups overlap almost completely.}
\label{tab:supp_conflation_ci}
\begin{tabular}{lccccc}
\toprule
Method & Cross & Retained & Regions & Pairwise IoU & Union \\
\midrule
ProtoPNet & 49.0 [47.8, 50.4] & 83.7 [82.3, 84.9] & 6.5 & 0.00 & 100.0 \\
ProtoPFormer & 33.9 [32.1, 35.7] & 83.7 [81.4, 85.8] & 9.3 & 0.00 & 41.3 \\
ProtoViT & 9.8 [8.3, 11.3] & 82.5 [81.7, 83.4] & 9.3 & 0.00 & 16.2 \\
PDiscoFormer & 42.8 [41.1, 44.5] & 66.0 [64.1, 67.9] & 7.9 & 0.00 & 22.6 \\
Zhu et al. & 20.7 [19.3, 22.2] & 85.6 [84.1, 87.1] & 5.0 & 0.00 & 46.2 \\
SOP & 1.7 [1.5, 1.9] & 11.6 [10.0, 13.4] & 20.0 & 0.97 & 21.1 \\
\midrule
Ours (leaky) & 20.4 [18.4, 22.5] & 86.0 [84.5, 87.5] & 5.2 & 0.00 & 33.2 \\
Ours & 0.0 [0.0, 0.0] & 5.9 [5.4, 6.3] & 5.2 & 0.00 & 33.2 \\
\bottomrule
\end{tabular}
\end{table*}

\Cref{tab:supp_conflation_ci} adds 95\% bootstrap intervals over images to the mean-colour columns of Tab.~2 and describes the regions that every method explains. Our model has exactly zero cross-part conflation on every image, since the evidence of a part is computed from its own patches alone, and the intervals of the leaky methods are far from zero. The low conflation of SOP comes with feature groups that overlap almost completely (mean pairwise IoU 0.97), so that removing the other groups removes few of the pixels of a group, and it does not imply that its groups explain separate regions.

\subsection{Pixel deletion and insertion}
\label{sec:supp_pixel}
\begin{table}[h]
\centering
\small
\caption{Pixel metrics on CUB (200 test images): deletion / insertion AUC normalised by the full-image confidence, Spearman stability of the map (\%). Ours: node terms over the fixed parts, each part's contribution spread evenly over its pixels.}
\label{tab:supp_pixel}
\begin{tabular}{lccc}
\toprule
Method & Deletion $\downarrow$ & Insertion $\uparrow$ & Stability $\uparrow$ \\
\midrule
Grad-CAM & \textbf{9.1} & 80.3 & 93.9 \\
PDiscoFormer & 16.6 & 86.2 & 92.1 \\
ProtoViT & 48.1 & 82.3 & 71.4 \\
Zhu et al. & 30.8 & 78.6 & 97.5 \\
ProtoPNet & 53.0 & 53.7 & 82.4 \\
ProtoPFormer & 30.5 & 88.4 & 91.5 \\
SOP & 16.1 & \textbf{93.8} & 96.7 \\
B-cos ViT & 14.7 & 36.9 & 83.4 \\
Ours & 17.2 & 86.2 & \textbf{99.4} \\
\bottomrule
\end{tabular}
\end{table}

\Cref{tab:supp_pixel} reports the pixel-level deletion and insertion metrics that are common for saliency maps, for which we spread the contribution of each of our parts evenly over its pixels. These metrics reward a map that concentrates on the few pixels to which the model is most sensitive, which favours Grad-CAM in deletion and SOP in insertion, whereas our explanation assigns one value to a whole part by design. Our maps are nevertheless competitive in both metrics and the most stable under input perturbations (99.4\%), since the parts and their contributions change little when the image is perturbed.

\subsection{Model randomisation and exactness}
\label{sec:supp_randomisation}
\begin{table}[h]
\centering
\small
\setlength{\tabcolsep}{2.5pt}
\caption{Model-randomisation test~\cite{adebayo2018sanity} (200 test images): rank correlation between the contrastive contributions of the trained and the randomised model, and accuracy (\%). Exactness: the terms reproduce the logit to a maximum absolute error of 1.5e-05 over all 5794 CUB test images (float32).}
\label{tab:supp_randomisation}
\begin{tabular}{lcccc}
\toprule
& \multicolumn{2}{c}{CUB} & \multicolumn{2}{c}{Aircraft} \\ \cmidrule(lr){2-3}\cmidrule(lr){4-5}
& Rank corr. & Acc. & Rank corr. & Acc. \\
\midrule
Trained model & 1.00 & 92.0 & 1.00 & 91.6 \\
Prototypes re-initialised & -0.00 & 1.5 & -0.00 & 1.5 \\
+ blocks re-initialised & -0.05 & 1.5 & 0.01 & 0.5 \\
\bottomrule
\end{tabular}
\end{table}

Following Adebayo \etal~\cite{adebayo2018sanity}, we re-initialise the class prototypes of the trained model and then also its fine-tuned blocks, and compare the contrastive contributions of every part and pair before and after. The rank correlation between the explanations of the trained and the randomised models drops from one to zero on both CUB and Aircraft while the accuracy drops to chance, so the explanations depend on what the model has learned. The node and pair terms reproduce the logit to a maximum absolute error of $1.5\times10^{-5}$ over all 5794 CUB test images, which is the rounding error of single-precision arithmetic.

\subsection{Part-swap test}
\label{sec:supp_swap}
\begin{table}[h]
\centering
\small
\setlength{\tabcolsep}{2.5pt}
\caption{Part-swap test (ViT-g, seed 0): one part is replaced by the same part of a test image of another class. Part term names the original / donor class, pair terms containing the part name the original class, image prediction stays original (\%), and the largest change of any other part's logit.}
\label{tab:supp_swap}
\begin{tabular}{lccccc}
\toprule
& \multicolumn{2}{c}{Swapped part term} & Pair terms & Image & Other parts \\ \cmidrule(lr){2-3}
Dataset & Original & Donor & Original & Original & max $|\Delta z|$ \\
\midrule
CUB & 0.3 & 66.7 & 38.4 & 81.8 & 0.028 \\
Aircraft & 0.3 & 75.3 & 41.2 & 80.9 & 0.038 \\
Cars & 0.2 & 65.5 & 40.2 & 81.2 & 0.037 \\
Flowers & 0.3 & 72.6 & 44.2 & 90.9 & 0.010 \\
Pets & 0.4 & 80.0 & 43.9 & 91.3 & 0.023 \\
Dogs & 0.5 & 57.1 & 35.4 & 82.5 & 0.044 \\
STL-10 & 2.8 & 80.4 & 48.9 & 68.5 & 0.022 \\
VOC & 12.5 & 71.5 & 55.0 & 48.7 & 0.031 \\
PIN & 0.4 & 68.8 & 42.7 & 82.6 & 0.036 \\
IN-S50 & 0.2 & 81.8 & 46.0 & 65.2 & 0.036 \\
\bottomrule
\end{tabular}
\end{table}

In the swap test, one part of a test image is replaced by the same part of a test image of another class, in its node stream and in every pair stream that contains it. On the fine-grained datasets of \cref{tab:supp_swap} the term of the swapped part names the original class in at most 0.5\% of the cases and the donor class in 57 to 80\% of them, while the term of every other part changes by at most 0.044, so a part's evidence comes from that part alone and the explanation can be edited one part at a time. The pair terms that contain the swapped part still name the original class in 35 to 44\% of the cases, as they legitimately see the other part of the pair, and the prediction of the whole image changes in 9 to 19\% of the cases depending on how distinctive a single part is for the dataset. On the general-object datasets the part terms follow the donor in 69 to 82\% of the cases, although on PASCAL VOC the swapped part still names the original class in 12.5\% of them, and the prediction of the image changes more often, since the classes of STL-10, PASCAL VOC and ImageNet-S50 differ in several parts at once.

\subsection{Leaky versus isolated model}
\label{sec:supp_leaky_vs_iso}
\begin{table*}[t]
\centering
\small
\setlength{\tabcolsep}{3.5pt}
\caption{Interpretability of the same model with full attention (leaky) and isolated (ours), ViT-g, 200 test images per dataset: conflation for three fills and part-level faithfulness measured by removing pixels.}
\label{tab:supp_leaky_vs_iso}
\begin{tabular}{llccccccccc}
\toprule
& & \multicolumn{2}{c}{Mean} & \multicolumn{2}{c}{Blur} & \multicolumn{2}{c}{Noise} & \multicolumn{3}{c}{Faithfulness (pixels)} \\ \cmidrule(lr){3-4}\cmidrule(lr){5-6}\cmidrule(lr){7-8}\cmidrule(lr){9-11}
Dataset & Model & Cross & Ret. & Cross & Ret. & Cross & Ret. & Rank corr. & Del. gain & Ins. gain \\
\midrule
CUB & Leaky (full attention) & 20.4 & 86.0 & 11.9 & 94.7 & 21.2 & 81.1 & 0.31 & 3.4 & 2.6 \\
 & Isolated (ours) & 0.0 & 5.9 & 0.0 & 18.8 & 0.0 & 6.2 & 0.90 & 9.2 & 9.5 \\
\midrule
FGVC-Aircraft & Leaky (full attention) & 20.9 & 79.1 & 13.3 & 92.1 & 22.0 & 76.5 & 0.37 & 2.2 & 2.0 \\
 & Isolated (ours) & 0.0 & 5.9 & 0.0 & 9.6 & 0.0 & 5.6 & 0.81 & 5.4 & 5.6 \\
\midrule
Stanford Cars & Leaky (full attention) & 12.9 & 77.9 & 8.8 & 92.3 & 13.3 & 76.1 & 0.60 & 3.8 & 3.2 \\
 & Isolated (ours) & 0.0 & 5.1 & 0.0 & 16.2 & 0.0 & 5.6 & 0.88 & 7.7 & 8.1 \\
\midrule
STL-10 & Leaky (full attention) & 3.2 & 95.5 & 1.2 & 99.8 & 2.7 & 94.9 & 0.27 & 0.7 & 0.6 \\
 & Isolated (ours) & 0.0 & 12.2 & 0.0 & 50.9 & 0.0 & 9.9 & 0.92 & 4.8 & 4.9 \\
\bottomrule
\end{tabular}
\end{table*}

\Cref{tab:supp_leaky_vs_iso} compares our model with the same model trained with full attention, with identical parts, prototypes, pair terms and losses. Removing pixels with the mean colour, a heavy blur or noise gives the same picture, as the leaky variant conflates 12 to 21\% of a part's evidence and keeps 81 to 95\% of it after the part's own pixels are removed, while our model has no conflation and keeps 6 to 19\%, which it reads from the replacement pixels of the part itself. Measured by removing the pixels of parts, the claimed contributions of the leaky variant predict the effect of the removal poorly (rank correlation 0.31) and those of our model well (0.90), so isolation, not the choice of parts, makes the explanation faithful. The rank correlation of 0.94 in Tab.~3 removes the tokens of a part instead of its pixels, which is the exact counterpart of how a part enters our model, and the pixel version is lower because the replacement pixels of a removed part are still encoded as that part's tokens, which keep part of its evidence (\cref{tab:supp_conflation_ci}), while in both protocols the parts are those of the original image and are not recomputed. The same comparison on FGVC-Aircraft, Stanford Cars and STL-10 (\cref{tab:supp_leaky_vs_iso}) gives the same picture on objects with very different structure, since our model has no conflation on any of them and keeps 5 to 16\% of a part's evidence after its pixels are replaced (51\% for a blurred part on STL-10, whose small images keep their colours under a blur), while the leaky variant conflates up to 22\% and keeps 76 to 100\%, and the claimed contributions of our parts predict the effect of their removal with a rank correlation of 0.81 to 0.92, against 0.27 to 0.60 for the leaky variant.

\subsection{Counterfactual edits}
\label{sec:supp_counterfactual}
\begin{figure*}[h]
  \centering
  \includegraphics[width=\linewidth]{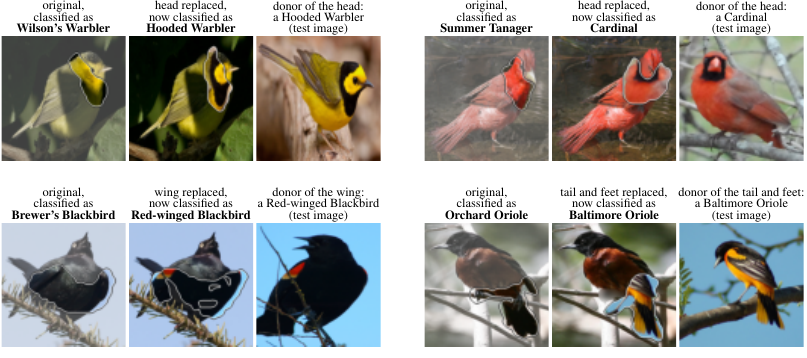}
  \caption{Counterfactual part swaps on CUB. Each example shows a correctly classified test image with one part outlined, the same image after that part has been replaced by the same part of a test image of a similar species, and the donor image. Replacing the part's tokens, in its node stream and in every pair stream that contains it, changes the prediction to the donor's species. The model swaps tokens, not pixels, and the edited image illustrates the swap by pasting the donor part's pixels at the position of the original part. The top row uses our main model with $K{=}6$ coarse parts (head), the bottom row the same model trained with $K{=}10$ parts, whose finer parts separate the wing (left) and the tail and feet (right) from the body. Donors were chosen among the test images of the target species with the most similar pose.}
  \label{fig:supp_counterfactual}
\end{figure*}
\begin{table}[h]
\centering
\footnotesize
\setlength{\tabcolsep}{3pt}
\caption{Counterfactual part swaps with the isolated model and its full-attention counterpart (CUB, ViT-g, node terms, same parts, prototypes and edits). One coarse part of a correctly classified test image is replaced by the same part of a test image of the runner-up class (1040 edits on 200 images). The explanation's prediction of an edit is the original explanation with the swapped part's term replaced by that part's term in the donor image.}
\label{tab:supp_cf_leaky_iso}
\begin{tabular}{lcc}
\toprule
 & Isolated & Full attn. \\
\midrule
Other parts' terms change (\% of a term) & 0.0 & 6.6 \\
Swapped part favours the donor (\%) & 64.7 & 11.3 \\
Swapped part favours the original (\%) & 3.1 & 86.3 \\
Prediction becomes the donor (\%) & 17.2 & 5.7 \\
Explanation predicts the decision (\%) & 100.0 & 84.4 \\
Error of predicted logit difference & $10^{-8}$ & 0.09 \\
Rank corr.\ of parts, predicted vs.\ actual & 1.00 & 0.52 \\
Most effective part predicted (\%) & 100.0 & 58.1 \\
\bottomrule
\end{tabular}
\end{table}

Since every term of our decision is computed only from its own part or pair, an explanation can be turned into a counterfactual by replacing a part and recomputing only the terms that contain it. On 60 correctly classified CUB test images, replacing a single part with the same part of an image of the runner-up class changes the prediction to that class in 14\% of the cases, most often for the part covering the breast and belly (60\%), and swapping parts in the order of their contribution reaches the runner-up for 97\% of the images, after a single part for 37 of them and after 1.4 parts on average, whereas removing a single part changes the prediction in only 2\% of the cases. \Cref{fig:supp_counterfactual} shows swaps between species that differ in one visible part, where a Brewer's blackbird that receives the wing of a red-winged blackbird, with its red shoulder patch, is classified as a red-winged blackbird, and an orchard oriole that receives the orange tail and feet of a Baltimore oriole is classified as a Baltimore oriole.

Isolation is what makes such an edit predictable from the explanation. \Cref{tab:supp_cf_leaky_iso} applies the same token-level swaps to our isolated model and to its full-attention counterpart, which uses the same parts and prototypes but lets every token attend to the whole image. In the isolated model the other parts' terms do not change at all, the swapped part takes on the evidence of the donor in 65\% of the edits, and the original explanation with the swapped term exchanged predicts the new decision exactly. In the full-attention model the swapped part is re-encoded in the context of the original bird, so it still favours the original species in 86\% of the edits and the prediction moves to the donor's species three times less often, while the other parts' terms change by 6.6\% of a term on average. The explanation then predicts the new decision in only 84\% of the edits and ranks the parts by their effect with a correlation of 0.52, so it does not tell the user which part would have to change for the decision to change.

\subsection{The effect of removing a part}
\label{sec:supp_deletion_identity}
The allocation $a_u = z_u + \tfrac12 \sum_{v} z_{uv}$ of part $u$ is not the same quantity as the change of the logit when the part is removed. Removing part $u$ drops its node term and the terms of all pairs that contain it, and since the pair terms are normalised by the number of pairs $|E|$, the remaining pairs are rescaled from $|E|$ to $|E'| = |E| - (K_u - 1)$, where $K_u$ is the number of parts of the image, while the node weights stay fixed. Because every remaining stream is encoded in isolation, none of the remaining raw terms changes, so the exact effect of the removal follows from the displayed terms alone,
\begin{equation}
  \Delta_u = z_u + \sum_{e \ni u} z_e - \Bigl(\frac{|E|}{|E'|} - 1\Bigr) \sum_{e \not\ni u} z_e ,
  \label{eq:supp_deletion}
\end{equation}
where $z_e$ are the pair terms as displayed. On 500 CUB test images, the effect predicted by \cref{eq:supp_deletion} matches the effect measured by a forward pass without the part to within 0.015 logits (mean error 0.002, against a mean effect of 2.7), which is the rounding error of bfloat16, with a rank correlation over the parts of an image of 0.9999. The allocation $a_u$ reaches 0.93 and the node term plus all incident pair terms 0.97, so the allocation ranks the parts almost as the removal does, while \cref{eq:supp_deletion} gives the removal effect itself. No such identity exists for a model with leakage, since removing a part changes the terms of all other parts.

\subsection{Shape and layout of the parts}
\label{sec:supp_shape}
Since the parts are computed from the whole image, their shape and position could carry class evidence of their own, which our guarantee would not cover. We therefore replace every pixel of every part by the mean colour, or by noise, while keeping the partition, so that only the shape and the layout of the parts remain. On 2000 CUB test images, the node terms of our model then classify 0.7\% of the images correctly with the mean colour and 0.5\% with noise (chance 0.5\%), against 37.5\% and 34.2\% for its full-attention counterpart, in which every token can still attend to the others. The isolated classifier thus draws no usable class evidence from the shape and layout of the parts, and the evidence it credits to a part comes from the part's pixels.

\subsection{Explanations of the ViT-B model}
\label{sec:supp_vitb_interp}
\begin{table}[h]
\centering
\footnotesize
\setlength{\tabcolsep}{3pt}
\caption{Our explanations with the ViT-B and the ViT-g backbone (CUB, same parts, 200 random test images unless stated). Conflation and retained evidence as in Tab.~2 (mean / blur / noise fill), part-level faithfulness as in Tab.~3 (token removal and pixel blanking), pixel deletion and insertion normalised by the full-image confidence (\cref{sec:supp_pixel}), and model randomisation (\cref{sec:supp_randomisation}), where heads and fine-tuned blocks are re-initialised.}
\label{tab:supp_vitb_interp}
\begin{tabular}{lcc}
\toprule
 & ViT-B & ViT-g \\
\midrule
Accuracy (\%) & 90.7 & 91.9 \\
\midrule
Conflation (\%) & 0 / 0 / 0 & 0 / 0 / 0 \\
Retained evidence (\%) & 9.6 / 30.3 / 8.7 & 5.9 / 18.8 / 6.2 \\
Shape-only acc.\ (\%, chance 0.5) & 0.8 & 0.7 \\
\midrule
Spearman, tokens & 0.91 & 0.94 \\
Del.\ / ins.\ gap, tokens (\%) & 15.3 / 17.1 & 15.7 / 16.7 \\
Spearman, pixels & 0.81 & 0.90 \\
Del.\ / ins.\ gap, pixels (\%) & 8.3 / 8.5 & 9.2 / 9.5 \\
\midrule
Pixel del.\,$\downarrow$ / ins.\,$\uparrow$ & 0.15 / 0.83 & 0.17 / 0.86 \\
Stability (\%) & 99.3 & 99.4 \\
\midrule
Randomised: Spearman & 0.01 & $-$0.05 \\
Randomised: acc.\ (\%) & 1.5 & 1.5 \\
\bottomrule
\end{tabular}
\end{table}

\Cref{tab:supp_vitb_interp} repeats the faithfulness evaluation for our ViT-B model, with the same parts. Cross-part conflation is exactly zero for every fill, as isolation guarantees for any backbone, and the part rankings remain faithful, with a rank correlation of 0.91 between the claimed contribution of a part and the effect of removing its tokens and deletion and insertion gaps of the same size as with ViT-g. Blanking the part's pixels gives a lower rank correlation than with ViT-g (0.81 against 0.90), and the retained evidence after blurring is higher (30.3 against 18.8\%), since a blurred part keeps its colours. Re-initialising the heads and the fine-tuned blocks destroys both the explanations and the accuracy for both backbones, so the faithfulness of our explanations does not depend on the larger backbone.

\subsection{Recomputing the mask and parts}
\label{sec:supp_recompute}
Our guarantee holds for the classifier given the partition. Writing $\Omega = g(x)$ for the crop, foreground mask and parts that are computed from the image $x$, every term of the decision is a function
\begin{equation}
z_u = h_u\big(x_{\Omega_u}, \Omega\big)
\label{eq:supp_conditional}
\end{equation}
of the pixels of its own part or pair and of the partition, which also sets the node weights and the normalisation of the pair terms, so that pixels outside $\Omega_u$ cannot change $z_u$ as long as $\Omega$ is unchanged, while $g$ itself sees the whole image and is not isolated. The experiments of \cref{sec:supp_conflation,sec:supp_pixel,sec:supp_randomisation,sec:supp_leaky_vs_iso} therefore keep the partition fixed, and \cref{tab:supp_recompute} measures what happens when it is recomputed after every edit, for our model and for its full-attention counterpart with the same parts and prototypes. When a part is replaced by the mean colour and the crop, mask and parts are recomputed from the edited image, the rank correlation between the claimed contribution of a part and the effect of its removal falls from 0.93 to 0.62 for our model, as the crop box moves in 42\% of the edits, whereas that of the counterpart falls from 0.37 to 0.06, so our explanation still predicts the effect of an edit when the partition is recomputed and the leaky one does not. When every pixel more than one patch away from the foreground is replaced, our part evidence does not change at all as long as the partition is kept, while that of the counterpart changes by 2.6\% for a mean-colour and by 14.3\% for a foreign background, whose evidence reaches the part tokens through attention and changes its prediction for 6.5\% of the images. Recomputed, both pipelines react to a foreign background in the same way, since it moves the crop box in 95\% of the images (box intersection over union 0.72) and changes the foreground mask even when the box is kept (intersection over union 0.73), so that our prediction changes for 21.5\% of the images against 18.5\% for the counterpart, and for 9.5\% when the box is kept, while a mean-colour background changes it for 1.5\%. The sensitivity to a foreign background thus lies in the unsupervised crop and foreground mask, which both models share, and not in the isolated classifier, and it is a limitation of the complete system that a guarantee conditional on the partition does not remove.

\begin{table}[h]
\centering
\footnotesize
\setlength{\tabcolsep}{3pt}
\caption{Recomputing the crop, foreground mask and parts after every edit (CUB, ViT-g, 200 correctly classified test images, node evidence as in \cref{tab:supp_leaky_vs_iso}). \emph{Part removal}: rank correlation between the claimed contribution of a part and the effect of replacing its pixels by the mean colour. \emph{Background}: every pixel more than one patch away from the foreground replaced by the mean colour or by the same pixels of another test image, with the change of the part evidence, $\sum_u |a'_u - a_u| / \sum_u |a_u|$ (\%), and the share of images whose prediction changes (\%). \emph{Fixed}, the original partition; \emph{box kept}, the original crop box with the mask and parts recomputed; \emph{recomputed}, the whole pipeline rerun on the edited image.}
\label{tab:supp_recompute}
\begin{tabular}{lcc}
\toprule
 & Ours & Full attention \\
\midrule
Part removal, fixed & 0.93 & 0.37 \\
Part removal, recomputed & 0.62 & 0.06 \\
\midrule
\multicolumn{3}{l}{\emph{fixed / box kept / recomputed}} \\
Mean background, evidence change & 0 / 6.6 / 12.3 & 2.6 / 4.6 / 6.2 \\
Mean background, prediction change & 0 / 0.5 / 1.5 & 0 / 0 / 1.0 \\
Foreign background, evidence change & 0 / 26.8 / 34.5 & 14.3 / 22.1 / 27.5 \\
Foreign background, prediction change & 0 / 9.5 / 21.5 & 6.5 / 11.0 / 18.5 \\
\bottomrule
\end{tabular}
\end{table}

\section{Accuracy}
\label{sec:supp_accuracy}

\subsection{Variation over seeds}
\label{sec:supp_seeds}
\begin{table*}[t]
\centering
\small
\setlength{\tabcolsep}{3.5pt}
\caption{Accuracy of our model as mean$_{\pm\mathrm{std}}$ over three seeds (superscript: number of seeds where fewer). Competitors are single runs.}
\label{tab:supp_seeds}
\begin{tabular}{lccccccccccc}
\toprule
Model & CUB & Aircraft & Cars & Flowers & Pets & NAB & Dogs & STL-10 & VOC & PIN & IN-S50 \\
\midrule
Ours (ViT-B) & 90.7$_{\pm0.3}$ & 90.3$_{\pm0.2}$ & 92.8$_{\pm0.1}$ & 99.0$_{\pm0.5}$ & 94.6$_{\pm0.2}$ & 91.6$^{1}$ & 87.7$_{\pm0.2}$ & 97.7$_{\pm0.1}$ & 93.3$_{\pm0.1}$ & 89.1$_{\pm0.1}$ & 96.7$_{\pm0.3}$ \\
Ours (ViT-g) & 91.9$_{\pm0.3}$ & 91.5$_{\pm0.2}$ & 93.3$_{\pm0.2}$ & 99.6$_{\pm0.2}$ & 96.3$_{\pm0.1}$ & 93.7$^{1}$ & 90.4$_{\pm0.3}$ & 98.6$_{\pm0.1}$ & 93.6$_{\pm0.5}$ & 90.7$_{\pm0.1}$ & 97.7$_{\pm0.1}$ \\
\bottomrule
\end{tabular}
\end{table*}

\Cref{tab:supp_seeds} gives the mean and standard deviation of our accuracies over three seeds, which change both the classifier's initialisation and the order of the training images while the parts stay fixed. The standard deviation is at most 0.5 points on every dataset, smaller than the margins over the competitors discussed in Sec.~4.4, and NABirds was run with one seed.

\subsection{Compute}
\label{sec:supp_compute}
\begin{table}[h]
\centering
\footnotesize
\setlength{\tabcolsep}{3pt}
\caption{Compute per test image (GFLOPs, exact count of one forward pass, CUB), grouped by backbone. \emph{Mask + parts}: our foreground mask (ViT-L, 1454) and part discovery (ViT-g, 5185), computed once per image and shared by both backbones. \emph{Classifier}: every part alone through the backbone and all pairs through the last blocks, for the image and its mirror image; for the other methods, their full forward pass.}
\label{tab:supp_compute}
\begin{tabular}{lrrrr}
\toprule
Method & Mask + parts & Classifier & Total & Params (M) \\
\midrule
\multicolumn{5}{l}{\textit{ViT-B}} \\
Black box (448) & -- & 215 & 215 & 87 \\
Zhu et al.\ (224) & -- & 51 & 51 & 94 \\
PDiscoFormer (518) & -- & 304 & 304 & 87 \\
SOP (448, 20 groups) & -- & 4513 & 4513 & 178 \\
Ours (448) & 6639 & 413 & 7052 & 87 \\
\midrule
\multicolumn{5}{l}{\textit{ViT-g}} \\
Black box (448) & -- & 2593 & 2593 & 1136 \\
Zhu et al.\ (224) & -- & 624 & 624 & 1165 \\
PDiscoFormer (518) & -- & 3578 & 3578 & 1137 \\
SOP (448, 20 groups) & -- & 54459 & 54459 & 2292 \\
Ours (448) & 6639 & 2456 & 9095 & 1136 \\
\bottomrule
\end{tabular}
\end{table}

The classifier itself costs less than one pass of the same backbone over the whole image (\cref{tab:supp_compute}, 2456 against 2593 GFLOPs with ViT-g), since each part passes through the frozen blocks alone with only its own tokens and only the pairs pass through the last blocks a second time, and it costs 22 times less than SOP, the other method that encodes its regions separately, with the same backbone. The additional cost of our model lies in the foreground mask and the part discovery, which run frozen backbones once over the image and its mirror image, and with them our ViT-g model costs 3.5 times a black-box pass of its backbone and a sixth of SOP with the same backbone.
The leaky part-based methods with the same backbone need a single pass over the image (624 GFLOPs for Zhu \etal\ and 3578 for PDiscoFormer), so the cost of our model lies between theirs and that of SOP, in exchange for explanations that leak nothing.

\begin{table}[h]
\centering
\small
\setlength{\tabcolsep}{3pt}
\caption{Cost of the isolated classifier against the number of coarse parts $K$ (ViT-g, last 4 blocks trained, RTX 3090, a foreground of 348 patches split evenly into $K$ parts as in an average CUB image, both flips). \emph{Frozen}: every part alone through the frozen blocks, computed once per image and cached during training. \emph{Trained}: all nodes and all $K(K{-}1)/2$ pairs through the last 4 blocks, GFLOPs and latency per test image. \emph{Step}: one optimisation step on 16 images, with peak memory. The foreground mask and the part discovery do not depend on $K$ (\cref{tab:supp_compute}).}
\label{tab:supp_k_scaling}
\begin{tabular}{lcrrrrr}
\toprule
& & \multicolumn{3}{c}{Inference} & \multicolumn{2}{c}{Training} \\ \cmidrule(lr){3-5}\cmidrule(lr){6-7}
$K$ & Pairs & Frozen & Trained & ms & Step (s) & GB \\
\midrule
2 & 1 & 1488 & 350 & 15 & 0.27 & 2.7 \\
3 & 3 & 1499 & 552 & 22 & 0.41 & 3.2 \\
4 & 6 & 1515 & 773 & 27 & 0.55 & 3.9 \\
6 (ours) & 15 & 1552 & 1280 & 45 & 0.99 & 5.3 \\
8 & 28 & 1574 & 1880 & 65 & 1.52 & 6.9 \\
10 & 45 & 1597 & 2581 & 98 & 2.23 & 8.7 \\
12 & 66 & 1669 & 3396 & 125 & 2.95 & 10.6 \\
\bottomrule
\end{tabular}
\end{table}

\Cref{tab:supp_k_scaling} shows how the classifier scales with the number of parts. The cost of the frozen blocks grows by only 12\% from $K{=}2$ to $K{=}12$, since the same foreground is split into more but smaller parts, which makes attention within a part cheaper while each part adds its own prefix tokens, whereas the trained blocks grow with the number of pairs, from 350 GFLOPs for $K{=}2$ to 3396 for $K{=}12$. Even at $K{=}12$, with 66 pairs, a training step on 16 images takes 3\,s and 10.6\,GB on one RTX 3090, so the quadratic number of pairs remains practical for the part counts at which parts are still meaningful (\cref{sec:supp_num_parts}).

\subsection{Input crop}
\label{sec:supp_crop}
\begin{table}[h]
\centering
\small
\setlength{\tabcolsep}{2.5pt}
\caption{Methods on CUB trained and tested on the full image and on our own mask crop (the crop our model uses), all other settings unchanged (final epoch, \%). ProtoViT's released hyper-parameters on CUB use ground-truth box crops (87.5); $^\ast$SOP is trained on our crop.}
\label{tab:supp_crop_competitors}
\begin{tabular}{lccc}
\toprule
Method & Full image & Our crop & $\Delta$ \\
\midrule
ProtoPNet & 82.9 & 84.0 & +1.1 \\
ProtoPFormer & 84.0 & 84.7 & +0.7 \\
ProtoViT & 81.7 & 87.1 & +5.4 \\
PDiscoFormer (our run) & 86.7 & 87.7 & +1.0 \\
Zhu et al. & 89.2 & 90.6 & +1.4 \\
B-cos ViT & 75.1 & 77.6 & +2.5 \\
SOP$^\ast$ & -- & 80.5 & -- \\
\midrule
Ours (ViT-g) & 91.6 & 91.9$_{\pm0.3}$ & +0.3 \\
\bottomrule
\end{tabular}
\end{table}

\begin{table}[h]
\centering
\small
\setlength{\tabcolsep}{2.5pt}
\caption{Our model on CUB with different input crops (ViT-g; parts restricted to the foreground mask computed in each frame). Accuracy (\%), mean$_{\pm\mathrm{std}}$ over three seeds.}
\label{tab:supp_crop_ours}
\begin{tabular}{lcc}
\toprule
Input & Full graph & Parts only \\
\midrule
No crop (full image) & 91.6$_{\pm0.1}$ & 90.7$_{\pm0.1}$ \\
Our mask crop (main) & 91.9$_{\pm0.3}$ & 90.8$_{\pm0.2}$ \\
Ground-truth box crop & 91.9$_{\pm0.1}$ & 90.7$_{\pm0.6}$ \\
\bottomrule
\end{tabular}
\end{table}

Our model sees the image cropped to its own foreground mask, which uses no annotation but removes most of the background. \Cref{tab:supp_crop_competitors} trains every competitor on the same crop, which raises their accuracy by 0.7 to 5.4 points, so the crop is a design choice that also helps the competitors and leaves our model ahead of each of them on CUB. Our own model gains only 0.3 points from the crop (\cref{tab:supp_crop_ours}), since the foreground mask already restricts its parts to the object and the crop mainly raises their resolution, and cropping to the annotated bounding box instead gives the same accuracy (91.9\%), so our own crop loses nothing against the annotation.

\section{Design choices}
\label{sec:supp_design}

\subsection{Number of parts}
\label{sec:supp_num_parts}
\begin{table}[h]
\centering
\small
\setlength{\tabcolsep}{3pt}
\caption{Number of coarse parts $K$ (CUB, ViT-g, 16 sub-parts). Test accuracy (\%) of the full graph and of the parts alone, as mean$_{\pm\mathrm{std}}$ over three seeds, and agreement of the coarse parts with the annotated keypoints (NMI / ARI).}
\label{tab:supp_num_parts}
\begin{tabular}{lcccc}
\toprule
$K$ & Full graph & Parts only & NMI $\uparrow$ & ARI $\uparrow$ \\
\midrule
2 & 92.3$_{\pm0.1}$ & 91.4$_{\pm0.2}$ & 31.8 & 11.2 \\
3 & 92.5$_{\pm0.1}$ & 91.7$_{\pm0.5}$ & 44.7 & 19.6 \\
4 & 92.2$_{\pm0.3}$ & 91.3$_{\pm0.1}$ & 49.1 & 20.9 \\
5 & 92.1$_{\pm0.5}$ & 91.3$_{\pm0.6}$ & 53.8 & 25.8 \\
6 (ours) & 91.9$_{\pm0.3}$ & 90.8$_{\pm0.2}$ & 56.8 & 29.2 \\
8 & 91.4$_{\pm0.2}$ & 90.5$_{\pm0.3}$ & 68.9 & 47.3 \\
10 & 91.1$_{\pm0.1}$ & 90.3$_{\pm0.3}$ & 67.0 & 44.5 \\
12 & 90.9$_{\pm0.1}$ & 90.0$_{\pm0.2}$ & 66.8 & 43.2 \\
\bottomrule
\end{tabular}
\end{table}

\begin{figure*}[h]
  \centering
  \includegraphics[width=\linewidth]{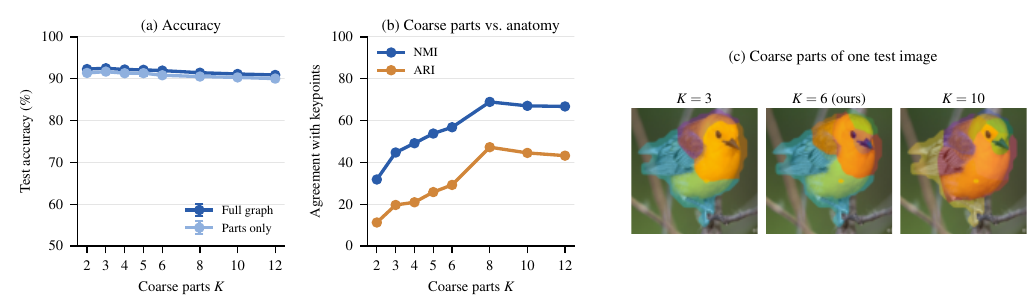}
  \caption{Number of coarse parts $K$ on CUB (ViT-g, 16 sub-parts). (a) Test accuracy, mean and standard deviation over three seeds. (b) Agreement of the coarse parts, which form the nodes of the graph, with the annotated keypoints (NMI / ARI). (c) The coarse parts of one test image.}
  \label{fig:supp_ksweep}
\end{figure*}
The number of coarse parts trades accuracy against the granularity of the explanation (\cref{tab:supp_num_parts,fig:supp_ksweep}). With two to five coarse parts every part covers a large share of the bird, and the accuracy is 0.2 to 0.6 points above our choice of $K=6$, but the parts agree poorly with the annotated anatomy (NMI 32 to 54), so that an explanation names regions such as the front and the back of the bird rather than its head, wings and breast. Beyond eight parts the agreement with the anatomy no longer improves while the accuracy keeps falling, as each isolated part becomes too small to be informative on its own. We use $K=6$, which keeps the accuracy within one point of the best value while dividing the bird into parts that align with its anatomy at the finer level (Tab.~5), and \cref{fig:supp_kgrid_cub} shows the parts for every $K$. Since the accuracy stays within two points over the whole range from two to twelve parts, $K$ can be chosen for the granularity of explanation a user needs, as with the ten parts of the counterfactual edits in \cref{fig:supp_counterfactual}, while the number of pairs, which grows as $K(K-1)/2$ and dominates the cost of the trained blocks (\cref{tab:supp_k_scaling}), makes six parts with 15 pairs a practical default.

\subsection{Hyper-parameter sensitivity}
\label{sec:supp_hparams}
\begin{table*}[t]
\centering
\small
\setlength{\tabcolsep}{3.5pt}
\caption{Hyper-parameter sweep (ViT-g, parts only, seed 0 and the mean with seed 1 where run). The sweep was run before the pipeline was fixed, with earlier part networks and the pad crop (with the annotated box on CUB), so its test accuracies are not those of Tab.~1 and serve only to compare the settings with each other. Stage 1 changes one setting of the initial configuration at a time and stage 2 combines the best; the setting we use (bold) has the highest \emph{held-out} accuracy averaged over the three datasets.}
\label{tab:supp_hparams}
\begin{tabular}{lccccccc}
\toprule
& & \multicolumn{2}{c}{CUB} & \multicolumn{2}{c}{Aircraft} & \multicolumn{2}{c}{Cars} \\ \cmidrule(lr){3-4}\cmidrule(lr){5-6}\cmidrule(lr){7-8}
Stage & Setting & Held-out & Test & Held-out & Test & Held-out & Test \\
\midrule
1 & lr 1e-05, head lr 1e-03, smoothing 0.1, wd 0.05, 10 ep., $\lambda$ 1.0 & 88.2 & 89.4 & 88.6 & 88.2 & 91.8 & 94.1 \\
1 & lr 1e-05, head lr 1e-03, smoothing 0.1, wd 0.05, 20 ep., $\lambda$ 0.5 & 88.0 & 89.0 & 88.9 & 88.4 & 92.6 & 94.2 \\
1 & lr 2e-05, head lr 1e-03, smoothing 0.1, wd 0.05, 10 ep., $\lambda$ 0.5 & 88.5 & 89.1 & 90.1 & 89.6 & 92.0 & 94.2 \\
1 & lr 4e-05, head lr 1e-03, smoothing 0.1, wd 0.05, 10 ep., $\lambda$ 0.5 & 89.0 & 89.5 & 90.7 & 90.4 & 92.3 & 94.2 \\
1 & lr 5e-06, head lr 1e-03, smoothing 0.1, wd 0.05, 10 ep., $\lambda$ 0.5 & 87.5 & 89.1 & 87.9 & 86.3 & 91.9 & 94.1 \\
1 & lr 1e-05, head lr 3e-03, smoothing 0.1, wd 0.05, 10 ep., $\lambda$ 0.5 & 88.3 & 89.2 & 89.3 & 88.7 & 91.9 & 94.0 \\
1 & lr 1e-05, head lr 1e-03, smoothing 0.0, wd 0.05, 10 ep., $\lambda$ 0.5 & 89.0 & 89.6 & 90.4 & 90.5 & 91.6 & 94.1 \\
1 & lr 1e-05, head lr 1e-03, smoothing 0.1, wd 0.2, 10 ep., $\lambda$ 0.5 & 88.3 & 89.6 & 89.7 & 89.3 & 92.4 & 94.3 \\
2 & lr 4e-05, head lr 1e-03, smoothing 0.0, wd 0.05, 10 ep., $\lambda$ 0.5 & 89.5 & 89.7 & 91.4 & 91.5 & 93.0 & 94.6 \\
2 & \textbf{lr 4e-05, head lr 1e-03, smoothing 0.0, wd 0.2, 10 ep., $\lambda$ 0.5} & 89.6 & 89.6 & 91.9 & 92.1 & 92.8 & 94.7 \\
2 & lr 8e-05, head lr 1e-03, smoothing 0.1, wd 0.05, 10 ep., $\lambda$ 0.5 & 89.3 & 89.5 & 91.9 & 92.2 & 92.5 & 94.7 \\
2 & lr 8e-05, head lr 1e-03, smoothing 0.0, wd 0.05, 10 ep., $\lambda$ 0.5 & 89.2 & 89.4 & 91.4 & 91.4 & 92.5 & 94.3 \\
\bottomrule
\end{tabular}
\end{table*}

The training settings of the classifier were chosen in two stages on held-out training images of CUB, Aircraft and Cars (\cref{tab:supp_hparams}), first changing one setting of an initial configuration at a time and then combining the best changes. The setting we use has the highest held-out accuracy averaged over the three datasets, and the test accuracy, shown for reference only, changes by less than one point across the second stage, so the results do not depend on a fine choice of the training settings.

\subsection{Parts against the ground truth}
\label{sec:supp_parts_gt}
\begin{table}[h]
\centering
\small
\caption{Part discovery against ground truth (CUB, all test keypoints): NMI / ARI between the part at each visible keypoint and the keypoint identity (background = its own label, as PDiscoNet). Flat discovery = the same discovery without the coarse-to-fine split. $^\dagger$from the PDiscoFormer paper.}
\label{tab:supp_nmi}
\begin{tabular}{lccc}
\toprule
Parts & $K$ & NMI $\uparrow$ & ARI $\uparrow$ \\
\midrule
Flat discovery & 6 & 60.0 & 31.5 \\
Refined, coarse level & 6 & 56.8 & 29.2 \\
Flat discovery & 16 & 60.4 & 35.0 \\
Refined, fine level & 16 & 68.6 & 51.0 \\
\midrule
PDiscoFormer$^\dagger$ & 8 & 69.9 & 43.5 \\
PDiscoFormer$^\dagger$ & 16 & 73.4 & 55.8 \\
\bottomrule
\end{tabular}
\end{table}

\Cref{tab:supp_nmi} measures how well the discovered parts match the 15 annotated keypoints of CUB with the metrics of PDiscoNet. Dividing the coarse parts into sub-parts raises the agreement at the finest level from 60.4 to 68.6 NMI and from 35.0 to 51.0 ARI compared with discovering 16 parts directly, so the hierarchy, not the larger number of parts, produces parts closer to the anatomy. PDiscoFormer, which learns its parts with a dedicated discriminative objective and full attention, agrees slightly better, while our parts are the most consistent and stable across images (Tab.~3), the properties that matter when a part is encoded on its own and compared across images. To see whether the parts depend on the initialisation of the discovery itself, we retrained it with two further seeds, which change the initial part centres and the episodes, and trained the classifier on each new partition with the same settings. Over the three discovery seeds the full graph reaches 92.0$\pm$0.2\% and the parts alone 90.6$\pm$0.3\%, and the agreement with the keypoints is 59.2$\pm$2.1 NMI and 33.0$\pm$3.4 ARI at the coarse level and 68.5$\pm$0.9 NMI and 50.9$\pm$1.9 ARI at the fine level, so the discovery finds parts of the same quality from every initialisation, and the coarse parts of our main model, which used the first seed, are if anything slightly less anatomical than those of the other two.

\subsection{What the pair terms compute}
\label{sec:supp_noxattn}
Each pair term of our model encodes its two parts in one stream, so its evidence could come from the joint appearance of the two parts or from interactions that attention computes between their tokens. To separate the two, we trained the same model with the two parts of every pair stream prevented from attending to each other. Each part then passes through the trained blocks exactly as in its own node stream, so the pair descriptor reduces to the token-weighted mean of the two node descriptors, which is still matched to the pair's own prototype after its own layer normalisation. This control reaches an accuracy comparable to that of our model with the same seeds, 91.8\% against 91.8\% on CUB and 91.3\% against 91.5\% on FGVC-Aircraft (two seeds each), while the parts alone reach 90.8\% on CUB (\cref{tab:supp_num_parts}). The relations that the pair terms add to the decision are therefore the joint appearance of two isolated parts, matched against a prototype for that pair of parts, rather than interactions computed by attention between them, which is consistent with each pair term seeing only the pixels of its own two parts. The control also shows that the pair terms could be computed from the node outputs without encoding the pairs a second time, which would remove the trained-block cost that grows with the number of pairs (\cref{tab:supp_k_scaling}).

\section{Human study}
\label{sec:supp_human}

We evaluated whether people could use explanations of CUB bird \cite{WahCUB_200_2011} classifications in a small, within-participant pilot. 33 participants each saw IsoPartFormer (our method), Grad-CAM \cite{selvaraju2017gradcam}, ProtoViT \cite{ma2024protovit}, and PDiscoFormer \cite{PDiscoFormer}. The study adapted HIVE’s \cite{kim2022hive} task-based evaluation framework. HIVE’s agreement task asks people to judge the correctness of a displayed model prediction; its distinction task asks them to identify the correct prediction among alternatives using visual explanations. Our protocol used 24 counterbalanced lists, with each participant assigned one list. Each list contained three trials per method in agreement and three per method in distinction, giving 12 trials per task and 24 trials per participant. In total, the design yielded 396 responses per task and 792 responses overall.

\subsection{Model Explanations}

Prior to beginning each task, participants were given a description of each model since prior AI experience was not a requirement for participation. The model was not named explicitly to avoid influencing their judgements through prior familiarity with, or expectations about, a particular method. The text and images were provided verbatim as follows:

\subsubsection{IsoPartFormer}
\textit{`This AI model makes predictions by identifying parts of an image and comparing them with the matching parts from an example image. The left shows the image being classified (The one that has been input by the user) with the parts identified by the model, whilst the right shows an example image of the class it predicted used for comparison. The matching colours and numbers highlight the corresponding parts in both images. The percentages show how much each part contributes to the model's prediction. Higher percentages mean that the part has a greater influence on the decision. Below is an example of a Red Footed Kittiwake explanation from Model 1 - Consider which parts drive the decision the most. Keep in mind that you should consider the percentages along with what part they correspond to, and if it makes sense for that decision.'}

\begin{figure}[h!]
  \centering
  \includegraphics[width=\linewidth]{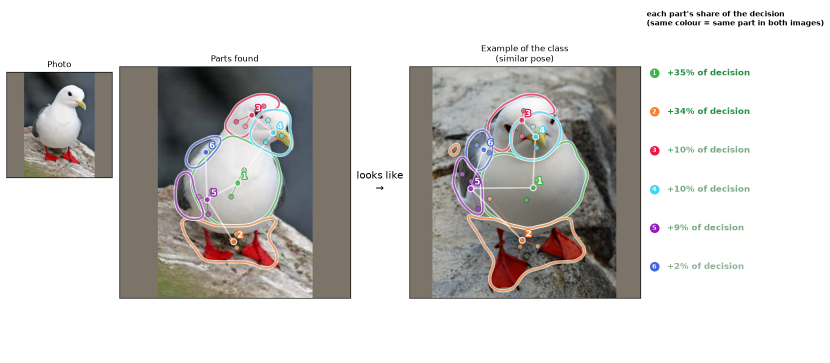}
  \caption{IsoPartFormer example given to participants within the study.}
\end{figure}

\subsubsection{PDiscoFormer}

\textit{`This AI model makes predictions by identifying which parts of an image provide the strongest evidence for a particular category. The left shows the image being classified, the middle highlights different regions identified by the model, and the right shows a heatmap indicating how strongly each region supports the prediction. The colours in the heatmap show the strength of the evidence. Red and yellow indicate stronger evidence, whilst blue and purple indicate weaker evidence. Below is an example of a Red Footed Kittiwake explanation from Model 3 - the model first identified the parts of the bird (seen in the middle image), and subsequently how those parts drive the decision with the most red parts being the most important.'}

\begin{figure}[h!]
  \centering
  \includegraphics[width=0.75\linewidth]{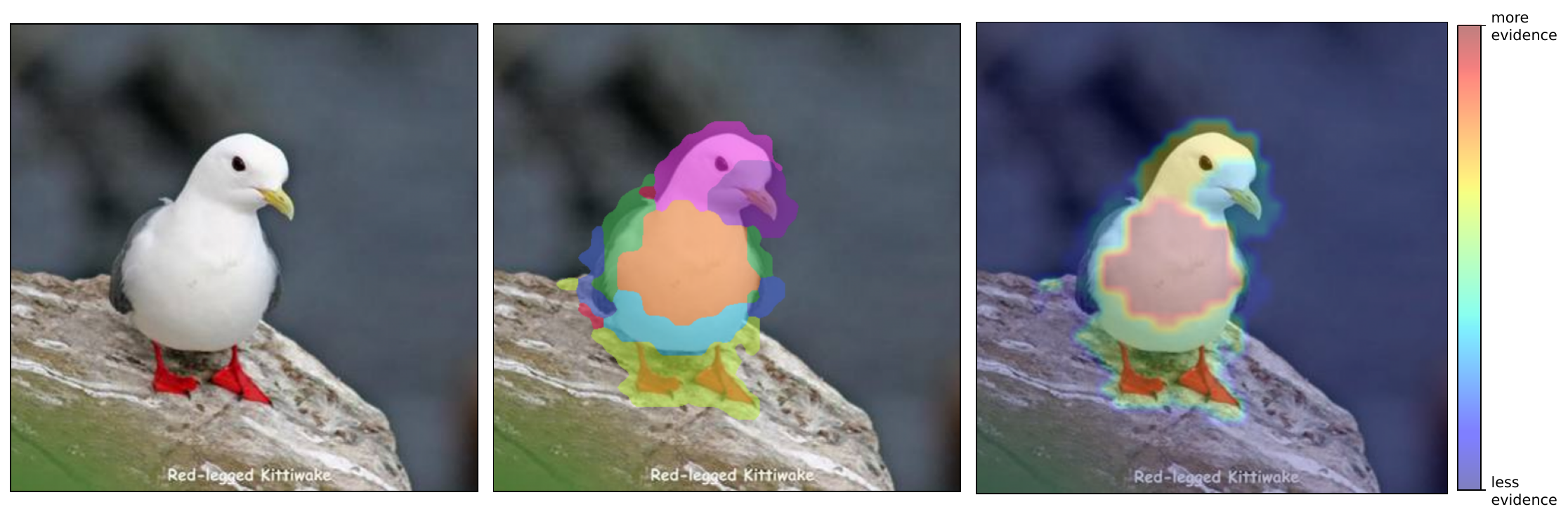}
  \caption{PDiscoFormer example given to participants within the study.}
\end{figure}

\subsubsection{Grad-CAM}

\textit{`This AI model makes predictions by analysing an image to recognise different categories. It uses heatmaps to show which parts of the image most strongly influenced each prediction. The left shows the image being classified. The colours in the heatmaps show how strongly each region contributed to the prediction. Red and yellow indicate stronger contributions, whilst blue and purple indicate weaker contributions. Below is an example of a Red Footed Kittiwake explanation from Model 4 - The areas that are shaded more red are what the model believes are the most important and salient to the decision.'} 

\begin{figure}[h!]
  \centering
  \includegraphics[width=0.75\linewidth]{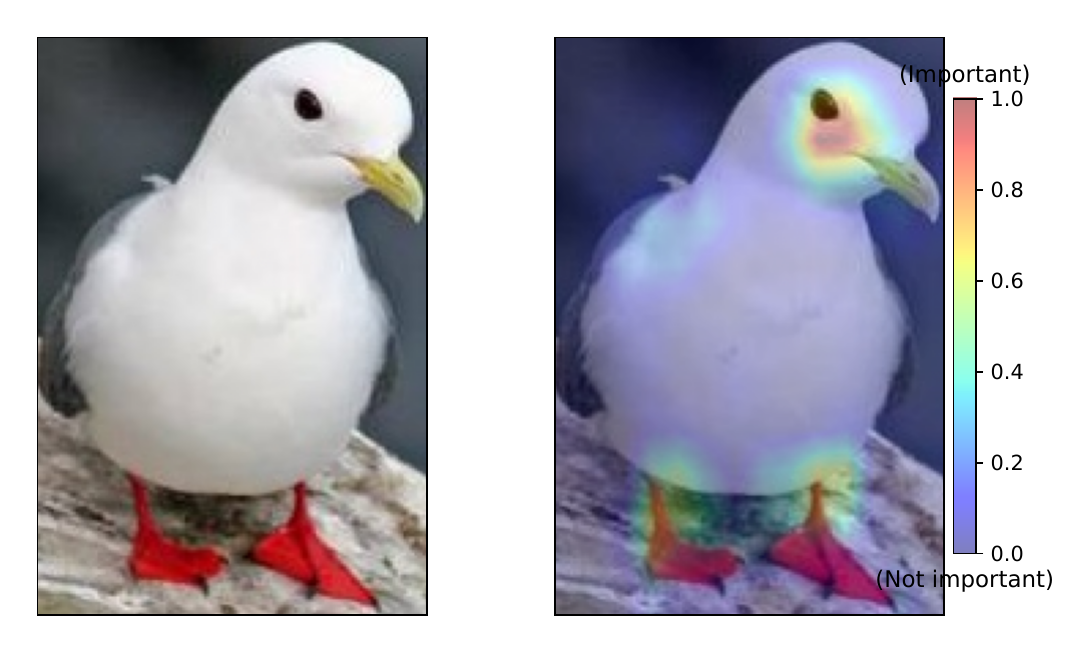}
  \caption{Grad-CAM example given to participants within the study.}
\end{figure}

\subsubsection{ProtoViT}

\textit{`This AI model makes predictions by comparing parts of an image with examples it has learnt from training on other images, called prototypes. The left shows the image being classified, whilst the right shows the prototypes the image is being compared with. The coloured squares highlight the matching regions in each image. The model uses these similarities to help decide which category the image belongs to. Below is an example of a Red Footed Kittiwake explanation from Model 2 - on the left is the original image with the squares indicating areas that matter to the model's decision, this is then compared with an image a similar bird the model chose with the squares showing the matching areas. The order of the images is random.'}

\begin{figure}[h!]
  \centering
  \includegraphics[width=0.5\linewidth]{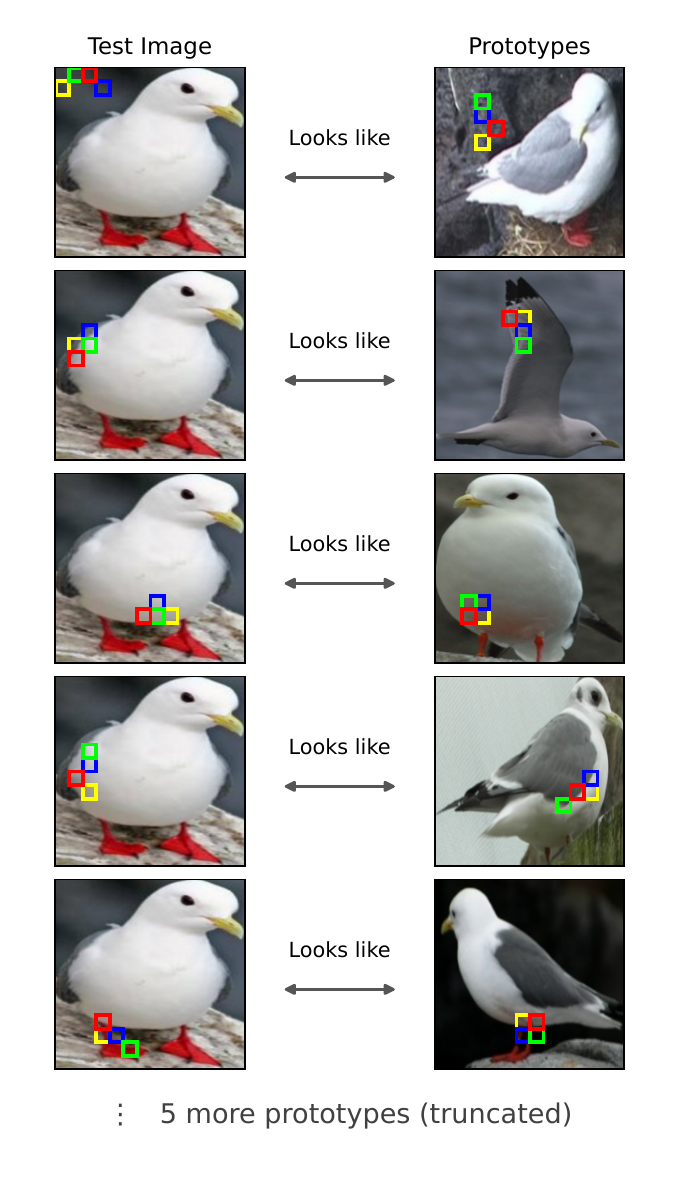}
  \caption{ProtoViT example given to participants within the study.}
\end{figure}

\subsection{Agreement Task}

The agreement task measured whether they could tell when the model’s prediction was correct. We report the hit rate on correct model predictions, the correct rejection rate on incorrect predictions, and their arithmetic mean as balanced accuracy. Prior to beginning the task, participants were shown the following explanatory text:

\textit{`In this task, you will be shown a series of bird images alongside an AI model’s explanation of how it arrived at a prediction, you will deliberately not be shown what the model predicted. Your task is to examine the image and the explanation provided by the model. Based on this information, you will be asked to judge how confident you are that the model's got it right. You don't need to know anything about birds. It's fine to be unsure, so just give your best judgement. You will be shown four different models in a random order, please read each model's description carefully.'}

\subsection{Distinction Task}

The distinction task measured whether participants could determine the correct prediction from a set of 4 options for each model. We report the fraction choosing the true class on all trials and separately on trials where the evaluated model predicted correctly or incorrectly. Prior to beginning the task, participants were shown the following explanatory text:

\textit{`In this task, you will be shown four explanations labelled A to D. Each one shows evidence the models finds for different possible species for the same input photo. We intentionally do not tell you the species, because we want you to focus on the explanations. Exactly one of the four is the bird's true species. The model can be wrong, so the true species isn't always the one the model favours. Your task is to examine the image and the explanations provided by the model. Based on this information, pick the explanation you think belongs to the true species, then say how confident you are. You will be shown four different models in a random order.'}

\subsection{Simulatability and confidence}

\textit{Simulatability} measures whether participants could anticipate the model’s decision from the explanation. For each trial, we compared the class selected by the participant with the class predicted by the model. A response counted as successful simulation when those classes matched, regardless of whether the model’s prediction was correct. We calculated simulatability as the percentage of such matches for each method. This measure differs from true class picked: a participant can identify the bird correctly while failing to predict an incorrect model decision, or predict the model’s incorrect decision without identifying the bird correctly.

Participants also rated how \textit{confident} they were in their selected answer, using a scale from ‘\textit{Not confident at all}’ to ‘\textit{Completely confident}’ over 4 options. We report mean confidence separately for answers that were correct and answers that were incorrect, in that order. For IsoPartFormer, these means were 2.95 and 2.32, respectively. The difference indicates that participants tended to express greater confidence when their answers were correct.

\subsection{Ethics and recruitment}

The pilot sample comprised 33 participants, with the study receiving approval from the Lancaster University FST Ethics Committee, as per the main paper. Participation was voluntary, and participants could withdraw at any stage of the study. Informed consent was collected prior to the completion of any tasks. No identifying information was collected, with the only subject-specific information being a self-identification of prior AI experience. A larger follow-up study is currently underway. 

\subsection{Analysis}

We treated the participant as the sampling unit, since each participant contributed repeated trials for every method. Uncertainty intervals were computed with a participant-cluster bootstrap of 2000 resamples, in which participants were drawn with replacement and all trials of a drawn participant were kept together, and the 2.5th and 97.5th percentiles of the resampled statistic give the 95\% interval. 

In each resample, the statistic was recomputed from the pooled trials of the drawn participants, and resamples in which a conditional rate was undefined, because no trial of that kind was drawn, were left out of the percentiles, while balanced accuracy then used the rate that was defined. This procedure captures the variation across participants while retaining the correlation among a person's responses. For the comparisons with IsoPartFormer, each participant's difference was the metric computed on their own trials with IsoPartFormer minus the same metric on their trials with the other method, and a sign-flip permutation test compared the absolute mean difference with its distribution under random sign flips of the participants' differences. With 33 participants the test was run with 20{,}000 random sign assignments, the $p$-values are two-sided without a plus-one correction, and no correction for multiple comparisons was applied. The reported $p$-values compare IsoPartFormer with each other method in the accuracy over all trials of the distinction task and in the balanced accuracy of the agreement task.

\subsection{Results}

Table \ref{tab:supp_human_full} shows that participants selected the true class most often with IsoPartFormer in the distinction task: 43.4\% of all trials, compared with 40.4\% for Grad-CAM, 39.4\% for ProtoViT and 25.3\% for PDiscoFormer. When the model’s prediction was correct, the rate for IsoPartFormer rose to 57.1\%, again the highest of the four methods. Its simulatability score was also highest at 51.5\%. The paired comparison for overall distinction performance gave \(p=0.02\) against PDiscoFormer, but \(p=0.81\) against Grad-CAM and \(p=0.65\) against ProtoViT. Mean confidence with IsoPartFormer was 2.95 for correct answers and 2.32 for incorrect answers, the largest separation shown in the table.
For the agreement task, Grad-CAM had the highest balanced accuracy (64.6\%) and AUROC (0.65), compared with 53.4\% and 0.57 for IsoPartFormer. PDiscoFormer had the highest hit rate (77.6\%) but the lowest correct rejection rate (24.0\%): participants judged its predictions correct on 76.8\% of trials. Overall, the pilot favoured IsoPartFormer for identifying the true class and anticipating its predictions, while Grad-CAM performed best when participants judged whether a displayed prediction was correct.

\begin{table*}[t]
\centering
\footnotesize
\setlength{\tabcolsep}{2.3pt}
\caption{Full results of the pilot human study on CUB ($N=33$, 99 trials per method and task). \emph{Distinction}: participants picked the true class among four candidates from the explanations, over all trials, over the trials on which the model was correct or wrong, how often they picked the model's prediction (simulatability), and their mean confidence on correct / incorrect answers (1 to 4). \emph{Agreement}: participants judged from one explanation whether the prediction was correct, with balanced accuracy, hit rate on correct predictions, correct rejection of wrong predictions, share judged correct and AUROC of the four-point answer. Brackets, 95\% bootstrap intervals over participants; $p$, sign-flip permutation test of the paired difference to ours.}
\label{tab:supp_human_full}
\begin{tabular}{l ccccccc @{\hspace{6pt}} cccccc}
\toprule
 & \multicolumn{7}{c}{Distinction} & \multicolumn{6}{c}{Agreement} \\
\cmidrule(lr){2-8}\cmidrule(lr){9-14}
Method & All $\uparrow$ & $p$ & Model correct $\uparrow$ & Model wrong & Simulat.\ $\uparrow$ & Confidence & & Bal.\ acc.\ $\uparrow$ & $p$ & Hit & Corr.\ rej. & Judged corr. & AUROC \\
\midrule
Grad-CAM & 40.4 {\scriptsize[31, 49]} & 0.81 & 49.0 {\scriptsize[36, 62]} & 32.0 & 38.4 & 2.70 / 2.34 & & \textbf{64.6} {\scriptsize[56, 73]} & 0.05 & 72.0 & \textbf{57.1} & 57.6 & \textbf{0.65} \\
ProtoViT & 39.4 {\scriptsize[31, 46]} & 0.65 & 54.0 {\scriptsize[42, 66]} & 24.5 & 46.5 & 2.51 / 2.42 & & 54.7 {\scriptsize[45, 65]} & 0.91 & 65.3 & 44.0 & 60.6 & 0.53 \\
PDiscoFormer & 25.3 {\scriptsize[18, 33]} & 0.02 & 30.0 {\scriptsize[18, 41]} & 20.4 & 33.3 & 2.40 / 2.54 & & 50.8 {\scriptsize[41, 60]} & 0.71 & 77.6 & 24.0 & 76.8 & 0.53 \\
\midrule
Ours & \textbf{43.4} {\scriptsize[33, 53]} & -- & \textbf{57.1} {\scriptsize[44, 70]} & 30.0 & \textbf{51.5} & 2.95 / 2.32 & & 53.4 {\scriptsize[44, 63]} & -- & 66.0 & 40.8 & 62.6 & 0.57 \\
\bottomrule
\end{tabular}
\end{table*}

\section{Further visualisations}
\label{sec:supp_vis}
\begin{figure*}[h!]
  \centering
  \includegraphics[width=\linewidth]{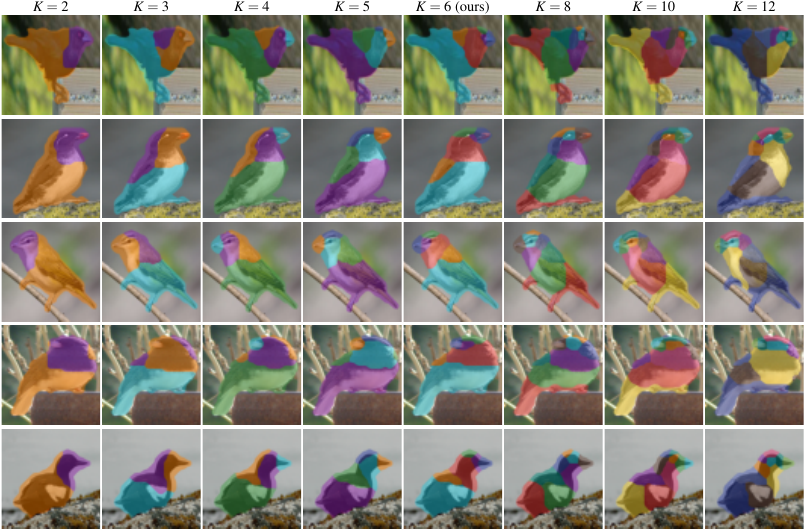}
  \caption{Coarse parts of CUB test images for every number of coarse parts $K$ (columns), each part shaded in a fixed colour per part index. As $K$ grows the bird is divided from front and back into head, breast, wings, tail and legs, and the parts follow the same regions across birds in different poses.}
  \label{fig:supp_kgrid_cub}
\end{figure*}
\subsection{Explanation examples}
\label{sec:supp_vis_expl}
\begin{figure*}[p]
  \centering
  \includegraphics[width=.92\linewidth]{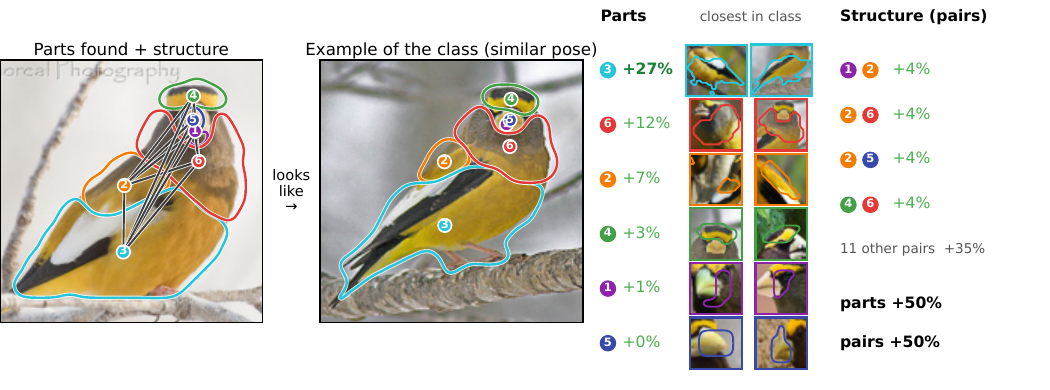}\\[4pt]
  \includegraphics[width=.92\linewidth]{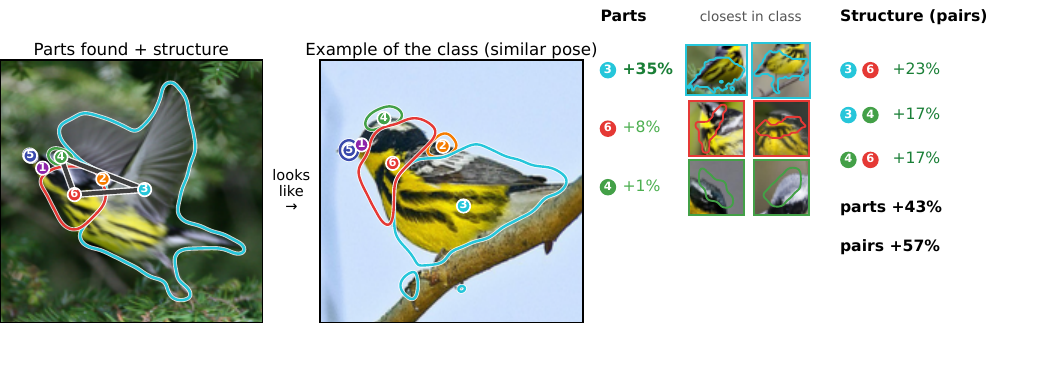}\\[4pt]
  \includegraphics[width=.92\linewidth]{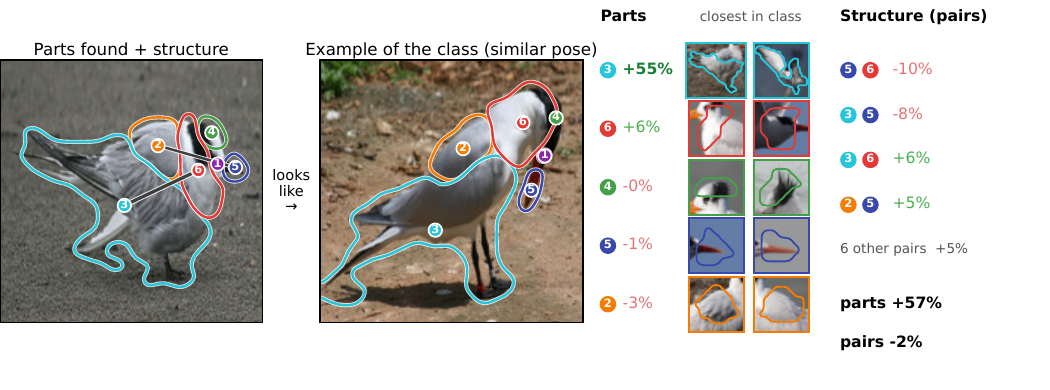}
  \caption{Explanations of our main model on CUB test images, an evening grosbeak and a magnolia warbler, both classified correctly, and a common tern classified as an elegant tern. Each row shows the parts and the pairs drawn as edges between part centres, the training image of the predicted class whose part layout is closest, the evidence of every part with the two training parts closest to it in the predicted class, and the evidence of the strongest pairs. Shares are the evidence of a term for the predicted class against the mean of the next three classes, relative to the summed absolute evidence of all terms, so the absolute shares of parts and pairs add up to 100\%.}
  \label{fig:supp_expl_cub}
\end{figure*}
\begin{figure*}[p]
  \centering
  \includegraphics[width=.92\linewidth]{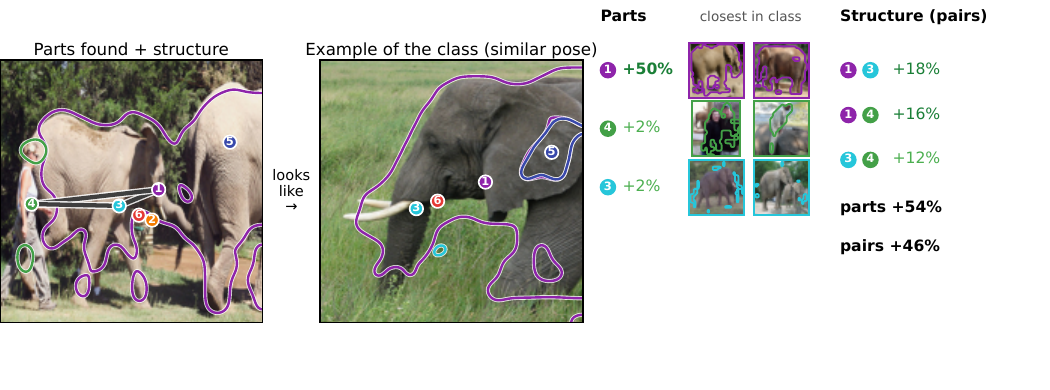}\\[4pt]
  \includegraphics[width=.92\linewidth]{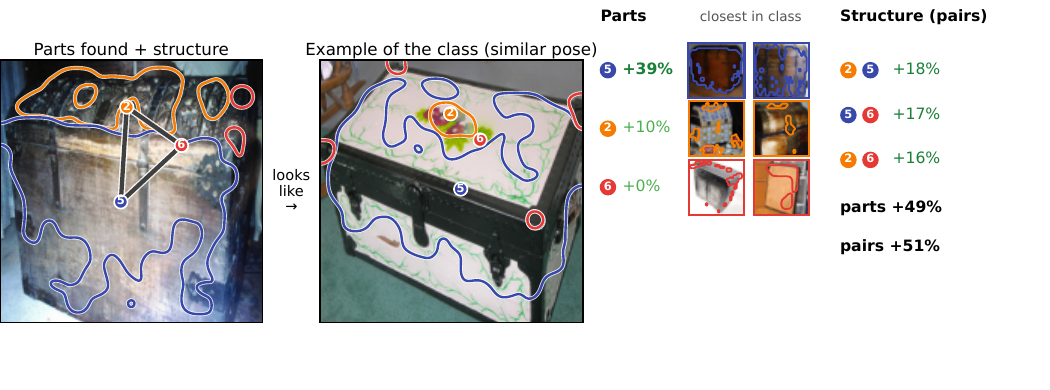}\\[4pt]
  \includegraphics[width=.92\linewidth]{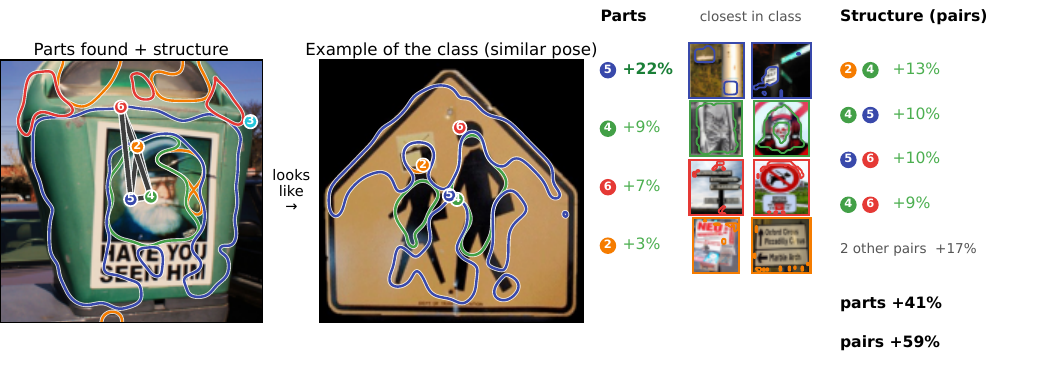}
  \caption{Explanations on ImageNet-S50, as in \cref{fig:supp_expl_cub}, for an African elephant and a chest, classified correctly, and a dustbin classified as a street sign.}
  \label{fig:supp_expl_ins50}
\end{figure*}
\begin{figure*}[p]
  \centering
  \includegraphics[width=.92\linewidth]{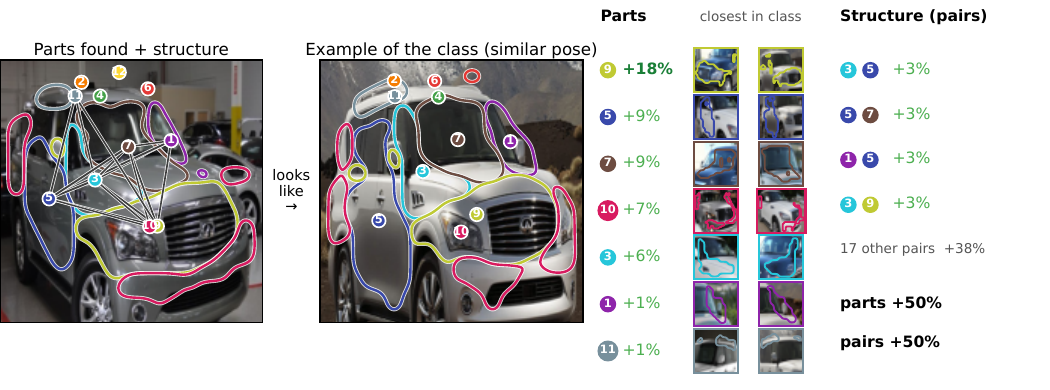}\\[4pt]
  \includegraphics[width=.92\linewidth]{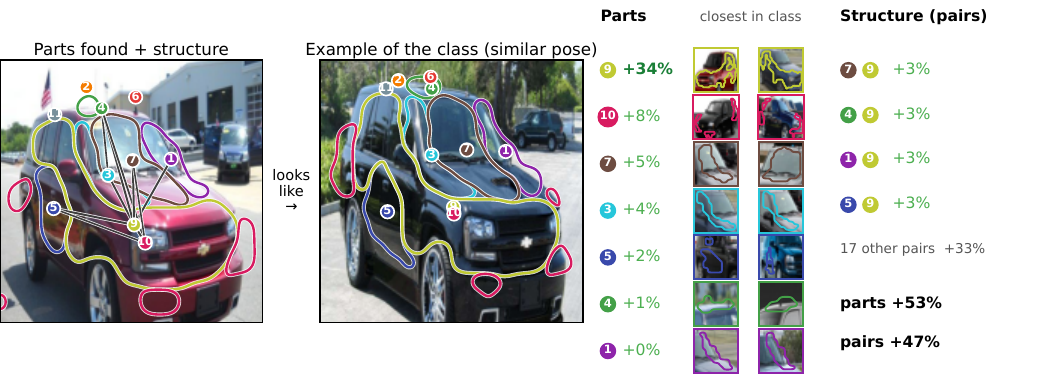}\\[4pt]
  \includegraphics[width=.92\linewidth]{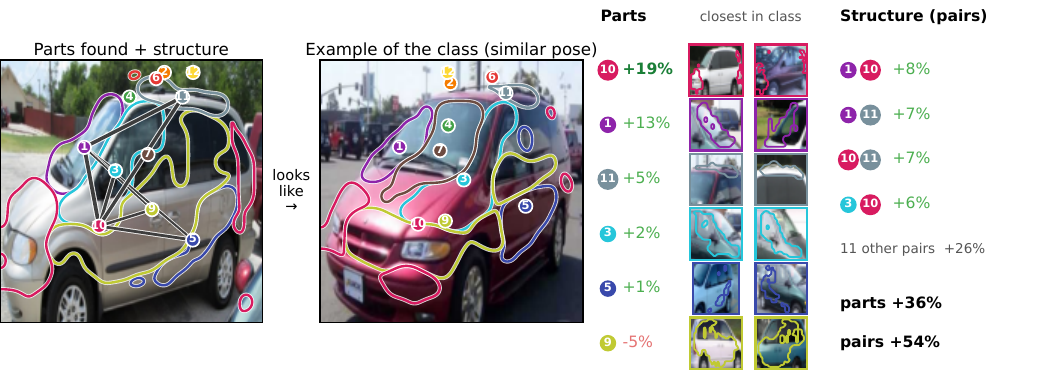}
  \caption{Explanations on Stanford Cars, as in \cref{fig:supp_expl_cub}, for an Infiniti QX56 and a Chevrolet TrailBlazer, classified correctly, and a Ram cargo van classified as a Dodge Caravan, with twelve coarse parts as in \cref{fig:supp_expl_aircraft}.}
  \label{fig:supp_expl_cars}
\end{figure*}
\begin{figure*}[p]
  \centering
  \includegraphics[width=.92\linewidth]{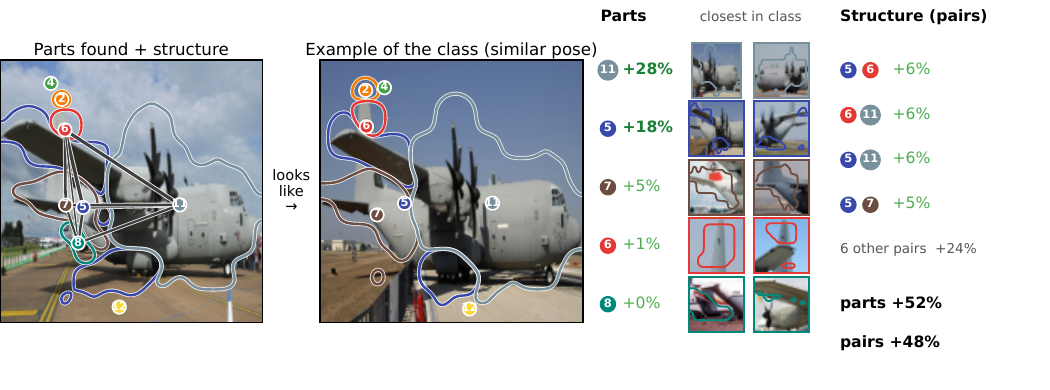}\\[4pt]
  \includegraphics[width=.92\linewidth]{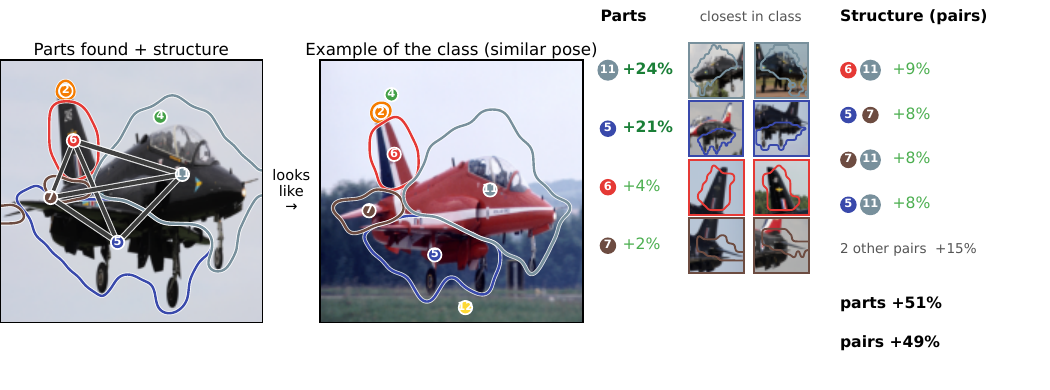}\\[4pt]
  \includegraphics[width=.92\linewidth]{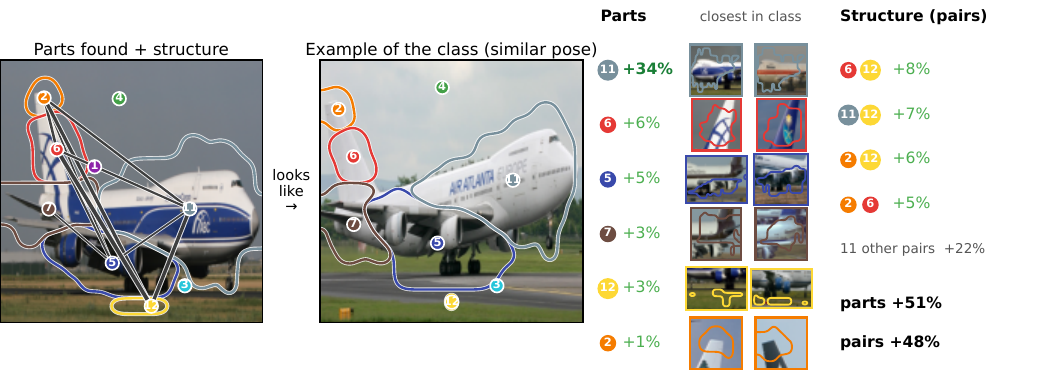}
  \caption{Explanations on FGVC-Aircraft, as in \cref{fig:supp_expl_cub}, for a C-130 and a Hawk T1, classified correctly, and a Boeing 747-300 classified as a 747-200, with twelve coarse parts, which divide the aircraft more finely than six.}
  \label{fig:supp_expl_aircraft}
\end{figure*}
\Cref{fig:supp_expl_cub,fig:supp_expl_ins50,fig:supp_expl_cars,fig:supp_expl_aircraft} show further explanations in the layout of the human study, each with a misclassified image. Every share is computed from the exact terms of the logits, so a reader can see not only which parts speak for the predicted class but also how much, and where a prediction goes wrong the explanation shows which region caused it. The common tern in \cref{fig:supp_expl_cub} is called an elegant tern mainly because of the appearance of its body, which holds more than half of the evidence, while the pairs, which add the joint appearance of neighbouring parts, speak slightly against that class. The 747-300 in \cref{fig:supp_expl_aircraft} is called a 747-200, a variant that differs mainly in the length of the upper deck, and the evidence comes mainly from the part that covers the front of the fuselage with its upper deck, while the two parts of the tail, which looks the same in both variants, add only 7\% together. The training parts next to each part show what the prototype of the predicted class looks like for that part, and they come from the class's own training images, since every part is encoded alone.

\subsection{Conflation examples}
\label{sec:supp_vis_conflation}
\begin{figure*}[h]
  \centering
  \includegraphics[width=\linewidth]{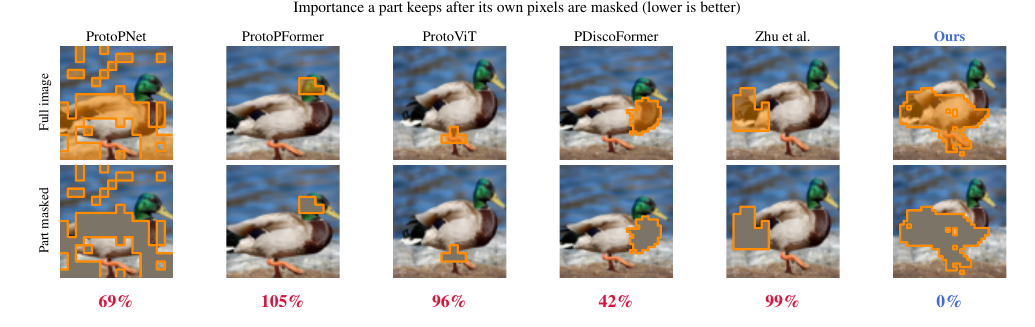}\\[2pt]
  \includegraphics[width=\linewidth]{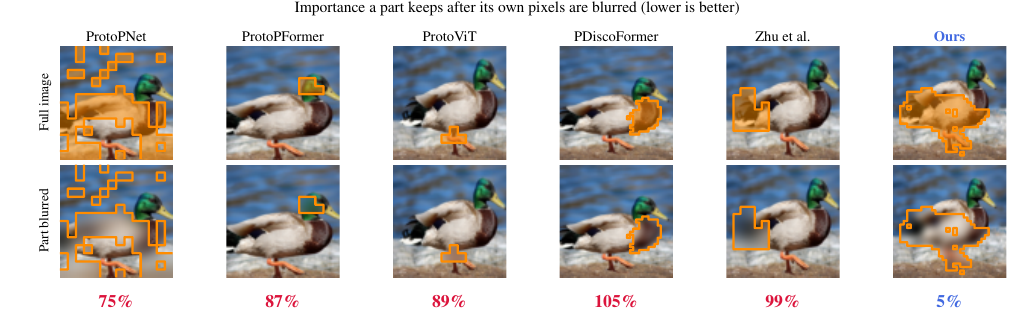}\\[2pt]
  \includegraphics[width=\linewidth]{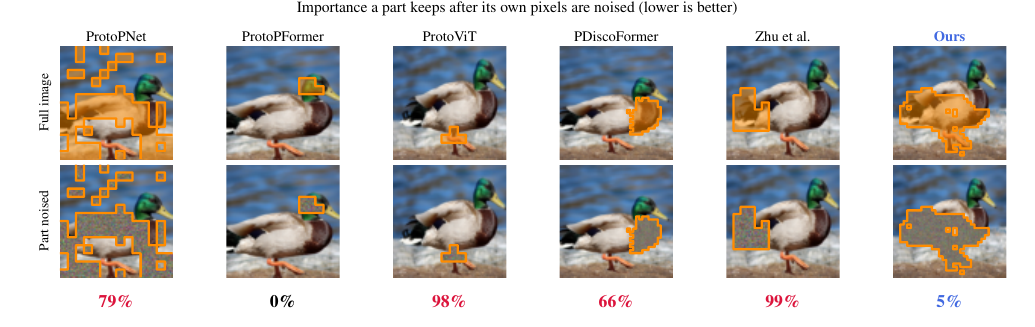}
  \caption{Retained evidence on one CUB test image with the three fills of Tab.~2 (mean colour, blur, noise). For every method, the most important explained region of the predicted class is outlined, its own pixels are replaced, and the share of its original importance that it keeps is given below (lower is better). Most competitors' regions keep most of their importance without their own pixels, since their tokens read the bird from the rest of the image, while our part keeps at most 5\% under every fill.}
  \label{fig:supp_retained_fills}
\end{figure*}
\begin{figure*}[h]
  \centering
  \includegraphics[width=\linewidth]{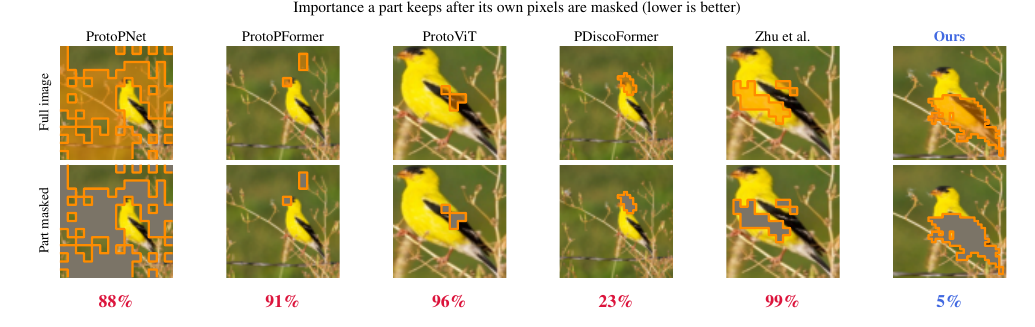}\\[2pt]
  \includegraphics[width=\linewidth]{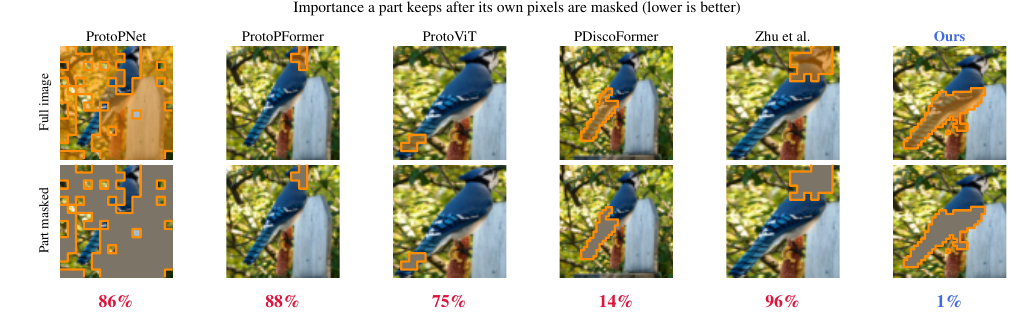}
  \caption{Retained evidence on two further CUB test images with the mean-colour fill, as in \cref{fig:supp_retained_fills}.}
  \label{fig:supp_retained_more}
\end{figure*}
\Cref{fig:supp_retained_fills,fig:supp_retained_more} illustrate the retained-evidence measure of Tab.~2 on single images. Masking the pixels of a competitor's most important region often leaves its importance almost unchanged, because the tokens of the region still receive the masked content from the rest of the image, while the importance of our part falls with its pixels, keeping at most 5\% in these examples.

\subsection{Parts for different numbers of parts}
\label{sec:supp_vis_parts}
\begin{figure}[h]
  \centering
  \includegraphics[width=\linewidth]{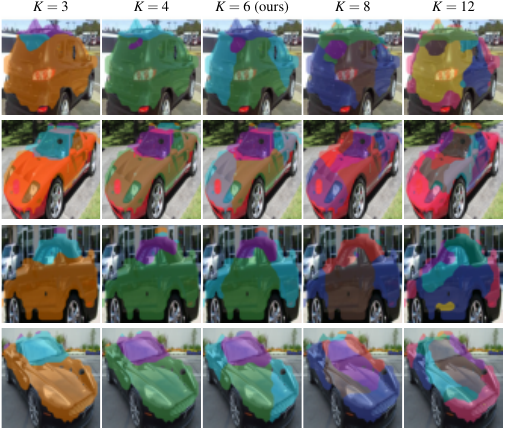}
  \caption{Coarse parts of Stanford Cars test images for different numbers of coarse parts $K$, as in \cref{fig:supp_kgrid_cub}.}
  \label{fig:supp_kgrid_cars}
\end{figure}
\begin{figure}[h]
  \centering
  \includegraphics[width=\linewidth]{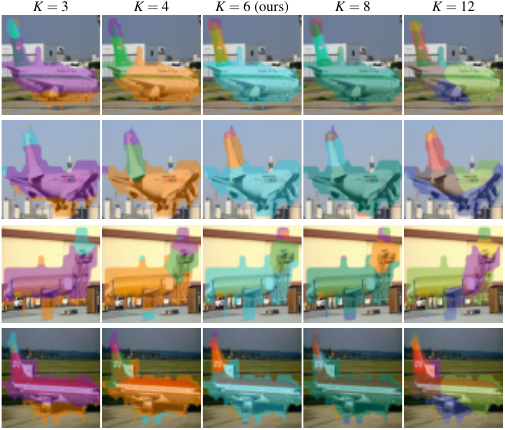}
  \caption{Coarse parts of FGVC-Aircraft test images for different numbers of coarse parts $K$, as in \cref{fig:supp_kgrid_cub}.}
  \label{fig:supp_kgrid_aircraft}
\end{figure}
\Cref{fig:supp_kgrid_cub} shows how the coarse parts of CUB change with the number of parts, on test images chosen for clean parts among 400 random correctly classified ones, and \cref{fig:supp_kgrid_cars} shows the same for Stanford Cars. On cars, three and four parts separate the windows and roof from the body, from six parts onwards the body is further split into its front and its side or rear, and more parts divide these again in similar places across cars of different shapes and viewpoints, while a small part takes the margin of background above the roof that the mask leaves. On aircraft, shown in \cref{fig:supp_kgrid_aircraft}, the fuselage and wings form one part from $K=6$ onwards, together with the margin of sky around them, while the remaining parts divide the tail, and only at three and four parts, and again at twelve, is the body itself split into a front and a rear part.

\subsection{Prototypes}
\label{sec:supp_vis_protos}
\begin{figure}[h]
  \centering
  \includegraphics[width=\linewidth]{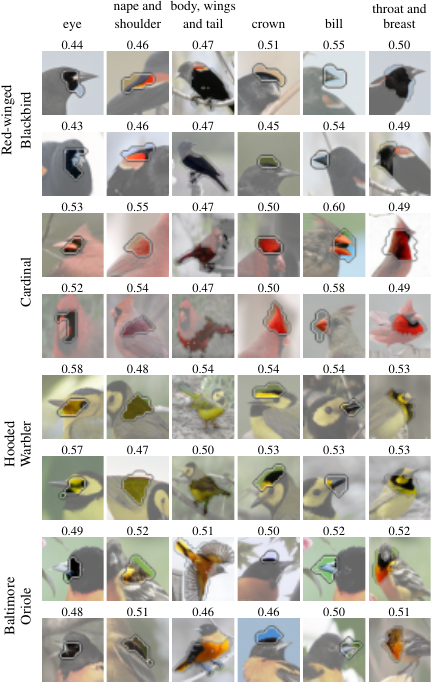}
  \caption{The node prototypes of our main CUB model, shown by the two training parts that are most similar to each prototype (cosine similarity after the model's layer normalisation, shown above each part, with every part encoded alone as at test time). Each row pair is one class and each column one of the six coarse parts, zoomed to the part. An asterisk would mark a neighbour from another class, and none of the parts shown is one.}
  \label{fig:supp_protos}
\end{figure}
Each class has one prototype per coarse part, and since every part is encoded in isolation, a prototype can be shown faithfully by the training parts closest to it, with no attention from the rest of the image contributing to the match. \Cref{fig:supp_protos} shows them for four classes, where the prototypes capture the part-specific marks of the species, such as the red shoulder patch of the red-winged blackbird in the shoulder part, the black hood of the hooded warbler in its crown and throat parts and the orange breast of the Baltimore oriole. Over all 200 classes and six parts, 99.2\% of the five training parts nearest to a prototype belong to the prototype's own class, and 96.2\% of the parts of the training images of a class are closer to their own class's prototype than to that of any other class, so every prototype is anchored in the parts of its own species.

\subsection{Fine sub-parts}
\label{sec:supp_vis_fine}
\begin{figure}[h]
  \centering
  \includegraphics[width=\linewidth]{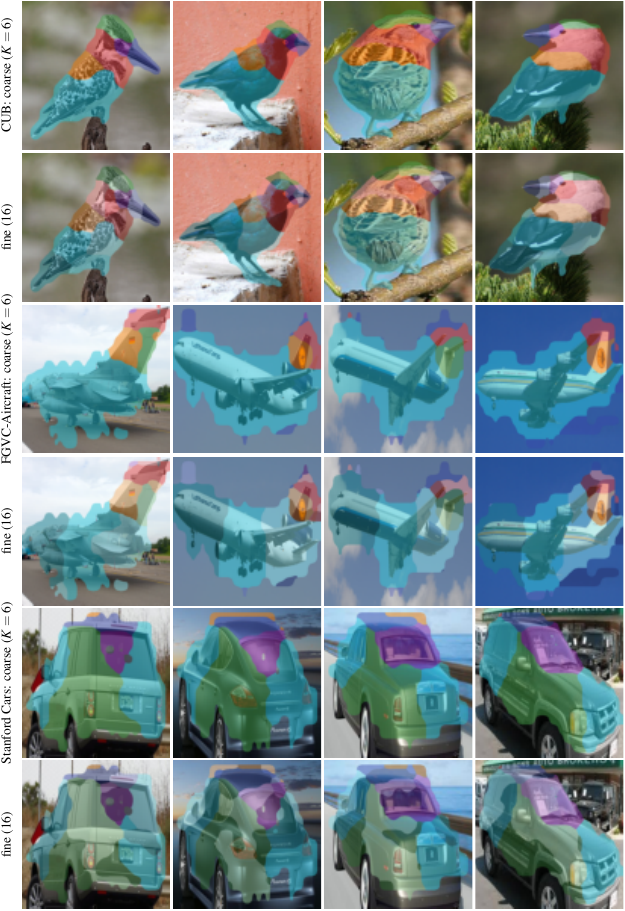}
  \caption{The six coarse parts of our model and their 16 fine sub-parts, each fine part shaded in a tint of the colour of its coarse parent, on test images of CUB, FGVC-Aircraft and Stanford Cars with at least four coarse parts that each hold 5\% of the object.}
  \label{fig:supp_levels}
\end{figure}
Our part discovery assigns every patch to one of 16 fine parts nested in the six coarse parts that the classifier uses (\cref{supp:discovery}). \Cref{fig:supp_levels} shows both levels. The fine parts divide each coarse part into smaller regions, such as the head and the body of a bird, the upper and the lower tail of an aircraft or the front and the side of a car, and since every fine part stays inside its parent, the two levels form a consistent hierarchy on every image.

\end{document}